%% file: main.tex
\documentclass[10pt,twocolumn,letterpaper]{article}

\usepackage[pagenumbers]{cvpr} 

\input{preamble}

\definecolor{cvprblue}{rgb}{0.21,0.49,0.74}
\usepackage[pagebackref,breaklinks,colorlinks,allcolors=cvprblue]{hyperref}

\def\paperID{***} 
\def\confName{CVPR}
\def\confYear{2026}

\title{One Language-Free Foundation Model Is Enough for Universal Vision \\ Anomaly Detection}

\author {Bin-Bin Gao\textsuperscript{a},
\quad Chengjie Wang\textsuperscript{a} \\
\normalsize 
\textsuperscript{a}{Tencent YouTu Lab} \\
\normalsize 
{\tt\small csgaobb@gmail.com}
}

\usepackage[bold]{hhtensor}
\usepackage{colortbl}
\usepackage{multirow}
\usepackage{wrapfig}
\usepackage{tabularx}
\usepackage{makecell}
\usepackage{enumitem}
\usepackage{marvosym}
\usepackage{pifont} 

\usepackage[accsupp]{axessibility} 
\usepackage{float} 
\usepackage{pifont}
\definecolor{lightgreen}{RGB}{240,255,255}
\definecolor{darkred}{RGB}{180,0,0} 
\definecolor{lightblue1}{RGB}{100, 149, 237} 
\definecolor{lightgreen1}{RGB}{34, 139, 34} 
\definecolor{lightorange1}{RGB}{255, 165, 0} 
\newcommand{\best}[1]{\textcolor{red}{\textbf{#1}}}
\newcommand{\sbest}[1]{\textcolor{blue}{\textbf{#1}}}

\newcommand{\gray}[1]{\textcolor{gray}{#1}}
\newcommand{\blue}[1]{\textcolor{blue}{#1}}
\newcommand{\green}[1]{\textcolor{green}{#1}}
\newcommand{\cmark}{\ding{51}} 
\newcommand{\xmark}{\ding{55}}
\newcommand{\pmerror}[2]{{\small #1}\tiny{\ensuremath{\pm}}{\scriptsize #2}}

\newcommand{\method}{UniADet}

\DeclareMathOperator*{\softmax}{Softmax}

\newcommand{\squishlist}{
 \begin{list}{$\bullet$}
  { \setlength{\itemsep}{0pt}
     \setlength{\parsep}{1pt}
     \setlength{\topsep}{1pt}
     \setlength{\partopsep}{0pt}
     \setlength{\leftmargin}{1.5em}
     \setlength{\labelwidth}{1em}
     \setlength{\labelsep}{0.5em} } }
\newcommand{\squishend}{
  \end{list}  }
  
\begin{document}
\maketitle

\input{sec/0_abstract}    
\input{sec/1_intro}
\input{sec/2_rework}
\input{sec/3_method}
\input{sec/4_exps}
\input{sec/5_cons}

\clearpage
{
    \small
    \bibliographystyle{ieeenat_fullname}
    \bibliography{main}
}

\input{sec/X_suppl}

\end{document}

%% file: preamble.tex
\newcommand{\red}[1]{{\color{red}#1}}



%% file: sec/0_abstract.tex
\begin{abstract}
Universal visual anomaly detection (AD) aims to identify anomaly images and segment anomaly regions towards open and dynamic scenarios, following zero- and few-shot paradigms without any dataset-specific fine-tuning. We have witnessed significant progress in widely use of visual-language foundational models in recent approaches.
However, current methods often struggle with complex prompt engineering, elaborate adaptation modules, and challenging training strategies, ultimately limiting their flexibility and generality.
To address these issues, this paper rethinks the fundamental mechanism behind visual-language models for AD and presents an embarrassingly simple, general, and effective framework for Universal vision Anomaly Detection (UniADet). Specifically, we first find language encoder is used to derive decision weights for anomaly classification and segmentation, and then demonstrate that it is unnecessary for universal AD. 
Second, we propose an embarrassingly simple method to completely decouple classification and segmentation, and decouple cross-level features, \ie, learning independent weights for different tasks and hierarchical features. 
UniADet is highly simple (learning only decoupled weights), parameter-efficient (only 0.002M learnable parameters), general (adapting a variety of foundation models), and effective (surpassing state-of-the-art zero-/few-shot by a large margin and even full-shot AD methods for the first time) on 14 real-world AD benchmarks covering both industrial and medical domains. We will make the code and model of \method~available at \url{https://github.com/gaobb/UniADet}.
\end{abstract}

%% file: sec/1_intro.tex
\section{Introduction}
\label{sec:intro}

\begin{figure}[htbp]
    \centering
    \includegraphics[width=1.0\columnwidth, keepaspectratio]{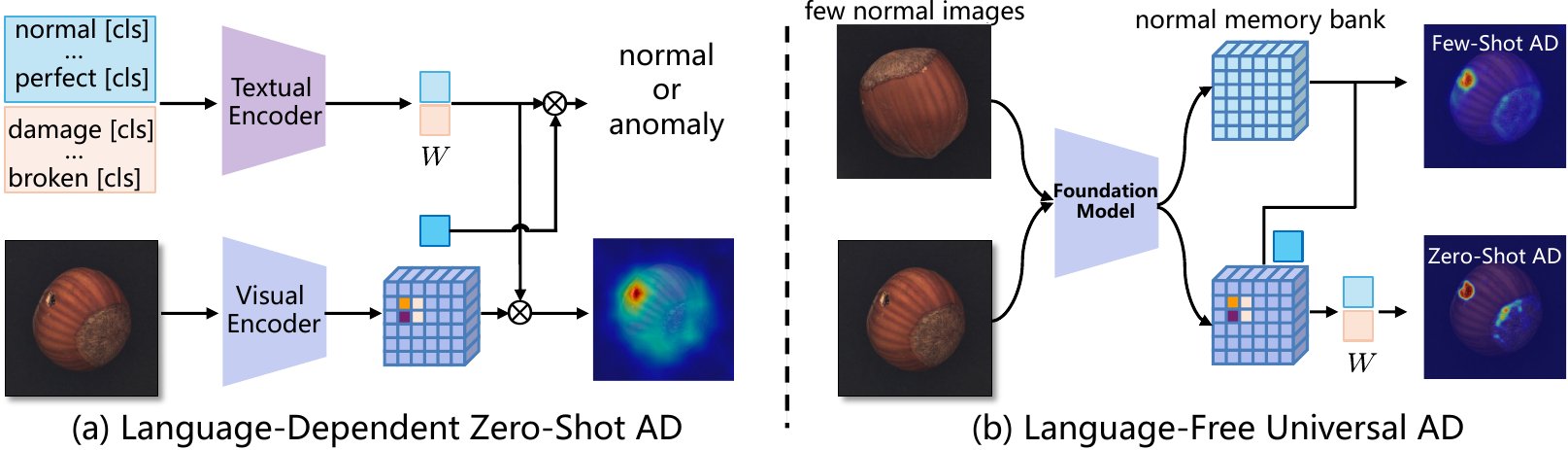} 
    \caption{\small{Comparisons of language-dependent zero-shot AD and our language-free universal AD (\textbf{\method}). UniADet is \emph{simple} (learning only task-related weights), \emph{parameter-efficient} (about 0.001M learnable parameters), \emph{general} (adapting a variety of foundation models), and \emph{effective} (surpassing state-of-the-art zero-/few-shot and even full-shot AD methods).
    }}
    \label{fig:zsadvsuniad}
    \vspace{-8pt}
\end{figure}

Visual Anomaly Detection (AD) is a foundational task in computer vision with broad applications, including surface defects detection in manufacturing~\cite{cvpr2019mvtec,eccv2022visa,cvpr2024realiad}, lesions identification in medical imaging~\cite{brainmri,Br35h,jhe2017endo,isic,colond}, and hybrid subspecies discovery in biological sciences~\cite {butterflyanomaly}. Traditional ADs~\cite{cvpr2022patchcore,neurips2022uniad,eccv2024onenip,dinomaly} typically require extensive training on specific normal images before they can detect anomalies. This paradigm, however, struggles with adaptability and scalability in open-world and privacy-sensitive scenarios. Recently, zero- or few-shot ADs have attracted more attention due to training-free on target domains, and have made significant progress, benefiting from powerful vision-language models (\eg, CLIP~\cite{clip}).

Past two years, a large number of CLIP-based zero-shot~\cite{cvpr2023winclip,iclr2024anomalyclip,eccv24adaclip,eccv24vcpclip,aaclip,bayes-pfl,adaptclip} and few-shot~\cite{cvpr2024inctrl,cvpr2024promptad} ADs have emerged and demonstrated impressive performance across domains. Nonetheless, these methods either become mired in complex prompt engineering, necessitate the design of elaborate adaptive modules, or rely on convoluted training strategies, as shown in Fig~\ref{fig:zsadvsuniad}(a). These issues inherently limit their flexibility and generality, particularly their ability to transfer to purely visual foundation models like DINOv3~\cite{dinov3}. This raises a crucial, unexplored question: Is a vision-language model really necessary for universal anomaly detection?

Recent MeatUAS~\cite{metauas} demonstrated that a pure vision model trained on synthetic images can achieve strong one-shot performance without vision-language models. Unfortunately, it cannot be extended to zero-shot scenarios. Furthermore, language should be independent of visual perception. Therefore, we want to explore how far universal AD (both zero- and few-shot) can go without language prompts or encoders, although visual-language-based methods are worth pursuing further.

We rethink vision-language-based ADs and find that both manually designed and learnable prompt embeddings derived from the text encoder serve as weights of image anomaly classification and pixel anomaly segmentation. We thus hypothesize that directly learning normal and anomaly embeddings should be equivalent to deriving embeddings from a text encoder. Therefore, it becomes possible to implement universal anomaly detection using a language-free foundation model, like DINOv2~\cite{dinov2r} and DINOv3~\cite{dinov3}. Furthermore, it is also possible to reduce model complexity and resource consumption.
In short, a novel, general, and language-free framework is present for universal anomaly detection.

Most existing vision-language-based AD methods share the weight embeddings on multi-level hierarchical features for image anomaly classification and pixel anomaly segmentation.
As we all know, deep vision representations extracted from different layers embed different manifolds. Furthermore, global image tokens and local patch tokens, even from the same layer, exhibit different distributions (see details in Fig.~\ref{fig:manifold}). Therefore, we believe that learning a shared weight to adapt these features of different manifolds will lead to conflicts, thereby reducing generalization. To mitigate this conflict, we propose a decoupled anomaly detector that decouples global anomaly classification and local anomaly segmentation with two independent weights. We further extend this decoupling principle across multi-level hierarchical features. 

Our main contributions are summarized as follows:
\squishlist
\item We rethink vision-language ADs and find that language prompts and encoders are unnecessary for zero-shot AD. This insight leads to an embarrassingly simple, parameter-efficient, and highly general language-free framework for universal anomaly detection.

\item We fully decouple global anomaly classification and local anomaly segmentation across multi-scale hierarchical features, effectively mitigating the learning conflict between different feature manifolds and substantially improving AD performance.

\item Comprehensive experiments covering 6 industrial and 8 medical benchmarks conclusively validate that our approach achieves state-of-the-art zero-shot and few-shot performance in both image-level classification and pixel-level segmentation. Notably, our few-shot UniADet is the first to outperform full-shot state-of-the-art.
\squishend

%% file: sec/2_rework.tex
\section{Related Works}
\label{sec:rework}

\noindent\textbf{Unsupervised AD} methods identify anomalies given a sufficiently normal training images. These methods can be broadly categorized into three main streams: embedding-based, discrimination-based, and reconstruction-based approaches. Embedding-based methods, such as PaDiM \cite{icpr2021padim}, PatchCore \cite{cvpr2022patchcore}, and flow-based models, CS-Flow \cite{wacv2022csflow} and PyramidFlow \cite{cvpr2023pyramidflow}, operate on the premise that features extracted offline from a pre-trained model maintain robust discriminative capabilities, thereby facilitating the separation of anomalies from normal samples in the feature space. Discrimination-based methods, including techniques like CutPaste \cite{cvpr2021cutpaste}, DRAEM \cite{iccv2021draem}, and SimpleNet \cite{cvpr2023simplenet}, typically reframe the unsupervised AD problem into a supervised one by synthesizing pseudo-anomalies for training. Reconstruction-based ADs, such as  RD~\cite{cvpr2022rd}, UniAD~\cite{neurips2022uniad}, OmniAL~\cite{cvpr2023omnial}, FOD~\cite{iccv2023focus}, OneNIP~\cite{eccv2024onenip}, and Dinomaly~\cite{dinomaly}, hypothesize that anomalous regions cannot be faithfully reconstructed due to their absence in the normal training distribution, thus leading to high reconstruction errors. However, they are inherently constrained to recognizing anomalies within seen classes during training and often exhibit poor generalization when applied to unseen (novel) classes. For a new application scenario, these models necessitate the expensive collection of sufficient normal images and subsequent retraining, which severely limits their rapid adaptability and practical applications.

\noindent\textbf{Zero-Shot AD} has recently gained significant traction by leveraging the powerful Vision-Language Models (VLMs), \eg, CLIP \cite{icml2021clip}. WinCLIP \cite{cvpr2023winclip} introduces a two-class textual prompt combined with multi-scale patch windows to achieve training-free AD.  
AnomalyCLIP \cite{iclr2024anomalyclip} learns class-agnostic prompt embeddings to facilitate patch-wise token alignment. AdaCLIP \cite{eccv24adaclip} and VCP-CLIP \cite{eccv24vcpclip} have built upon similar concepts, further integrating visual knowledge into the textual prompt embeddings. Bayes-PFL~\cite{bayes-pfl} models the prompt space as a learnable probability distribution from a Bayesian perspective. AA-CLIP~\cite{aaclip} proposes a two-stage approach that first learns prompt embeddings and then aligns vision features. FAPrompt~\cite{faprompt} learn a set of complementary, decomposed abnormality prompts.
AdaptCLIP~\cite{adaptclip} treats CLIP as a foundational service, proposing alternating and comparative learning strategies based on three lightweight adapters.
We contend that these additional modifications introduce unnecessary model complexity and potentially degrade the robust, inherent representational capabilities of the original CLIP model. 
Alternative approaches like ACR~\cite{nips2023acr} and MuSc \cite{iclr2024musc} achieve zero-shot AD solely by exploiting batch-level or full-shot testing image information, but it may be impractical in privacy-sensitive deployment environments. 

\noindent\textbf{Few-Shot AD.} Few-Shot AD focuses on model learning or adaptation using only a highly limited number of nominal training images. Early efforts, including TDG \cite{iccv2021tdg}, RegAD \cite{eccv2022regad}, and FastRecon \cite{iccv2023fastrecon}, have generally shown performance lagging behind their unsupervised counterparts. A separate few-shot setting also considers limited anomalous examples \cite{cvpr2022catching, cvpr2023bgad}. The integration of VLMs has recently spurred significant improvements. WinCLIP+ \cite{cvpr2023winclip} pioneered the application of CLIP to few-shot AD by storing normal tokens in a memory bank, retrieving the nearest token for each query via cosine similarity, and computing the anomaly map based on the nearest distance. InCtrl \cite{cvpr2024inctrl} further enhanced this by holistically integrating multi-level information for image-level anomaly classification, though it neglects pixel-level segmentation. PromptAD \cite{cvpr2024promptad} introduced the concept of an explicit anomaly margin to mitigate the training instability caused by the lack of anomalous training images. However, this method mandates the computationally expensive re-training of the model for each new target dataset. Recently, UniVAD~\cite{gu2025univad} proposes a training-free method for detecting cross-domain anomalies. This method combines multiple pre-trained models with different functions, such as RAM, Grounding DINO, SAM, CLIP, and DINOv2~\cite{dinov2r}, and achieves excellent few-shot AD performance. In contrast, we aim to explore few-shot AD performance in a training-free and computationally efficient manner using only a single foundation model.

%% file: sec/3_method.tex
\section{Methods}\label{sec:methods}
\noindent\textbf{Problem Formulation}:
Our objective is to develop a universal anomaly detection framework capable of detecting anomalies across diverse domains without any training or fine-tuning on the target dataset. This approach inherently assumes a significant distributional shift between the training and testing domains. Formally, let $\mathcal{D}_\text{base} =\{X_i, Y_i, y_i\}_{i=1}^N$ be the base training dataset, comprising $N$ images, $X_i \in \mathcal{R}^{H \times W \times 3}$ denotes the $i$-th image, and $Y_i \in \mathcal{R}^{H \times W}$ and $y_i \in \{0, 1\}$ are the corresponding anomaly mask and image-level label, respectively, where $y_i=0$ indicates normal and $y_i=1$ signifies an anomaly. The testing set $\mathcal{T}$, consists of multiple novel domains $\mathcal{D}_\text{novel}^t=\{X_i, Y_i, y_i\}_{i=1}^{N_t}$, featuring distinct object categories and anomaly types. In zero-shot setting, the model must perform both image-level anomaly classification and pixel-level anomaly segmentation on $\mathcal{D}_\text{novel}^t$ without accessing any data from this domain during the training phase. In few-shot setting, a few normal images ($k$), $\mathcal{P}_c = \{X_i\}_{i=1}^k$, is available for each target class $c$ (typically $k \in \{1, 2, 4\}$).

\noindent\textbf{Overview}: 
\begin{figure}[htbp]
    \centering
    \includegraphics[width=1.0\columnwidth, keepaspectratio]{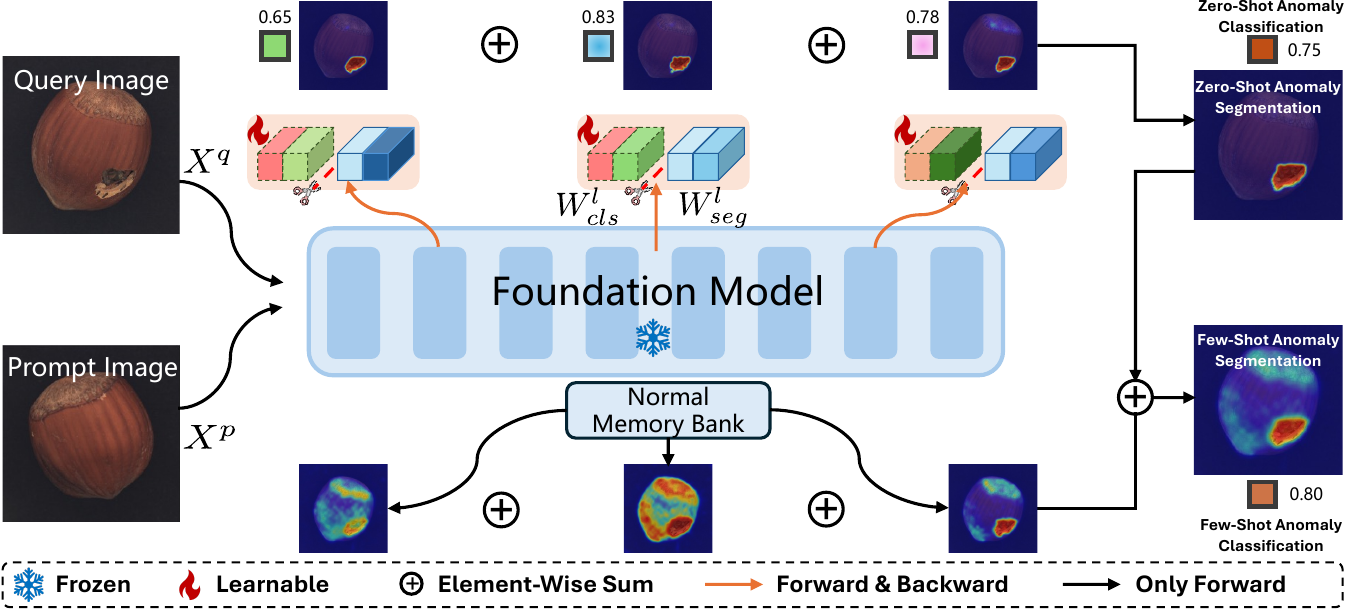} 
    \vspace{-12pt}
    \caption{\small{The language-free \textbf{\method}~framework. UniADet decouples global image anomaly classification and local patch segmentation so that their weights are learned independently across hierarchical features. Once trained, it can identify any anomaly images and segment anomaly regions, providing only few-shot and even zero-shot normal images.
    }}
    \label{fig:uniadet-framework}
    \vspace{-5pt}
\end{figure}
As illustrated in Fig.~\ref{fig:uniadet-framework}, the proposed UniADet consists of Decoupling Classification and Segmentation (Sec.~\ref{subsec:dcs}) and Decoupling Hierarchical Features (Sec.~\ref{subsec:dhf}). Below, we present them in detail.

\subsection{Rethinking VLMs for Anomaly Detection}\label{sec:revlmad}
\begin{table}[t]
	\caption{Components comparison of state-of-the-art Zero-Shot (ZS) and Few-Shot (FS) AD methods. \red{\cmark} indicates used or satisfied, \green{\xmark} indicates not used or satisfied.}\label{tab:rethinksota}
	\centering
	\label{Tab1}
    \vspace{-8pt}
	\renewcommand{\arraystretch}{1.0}
	\resizebox{1\columnwidth}{!}
	{
		\begin{tabular}{c|ccccc c|cc}
			\toprule
			Methods	&$\mathcal{F(\cdot)}$ &$\vec \theta_v$ &$\mathcal{T(\cdot)}$  &$\vec \theta_t$  & $T_f$    & $T_l$    &ZS &FS           \\  
            \midrule
            WinCLIP~\cite{cvpr2023winclip}         &\red{\cmark}    &\green{\xmark} &\red{\cmark}  &\green{\xmark}   &\red{\cmark}    &\green{\xmark}  &\red{\cmark}     &\red{\cmark}  \\
            AnomalyCLIP~\cite{iclr2024anomalyclip} &\red{\cmark} &\green{\xmark}  &\red{\cmark}  &\red{\cmark}     &\green{\xmark}  &\red{\cmark}    &\red{\cmark}     &\green{\xmark}  \\
            AdaCLIP~\cite{eccv24adaclip}       &\red{\cmark}   &\red{\cmark}  &\red{\cmark}   &\red{\cmark}     &\red{\cmark}   &\green{\xmark}      &\red{\cmark}     &\green{\xmark}  \\
            InCtrl~\cite{cvpr2024inctrl}        &\red{\cmark}   &\red{\cmark}  &\red{\cmark}   &\red{\cmark}     &\red{\cmark}    &\green{\xmark}    &\green{\xmark}     &\red{\cmark} \\
            
            Bayes-PFL~\cite{bayes-pfl} &\red{\cmark}   &\red{\cmark}  &\red{\cmark}   &\green{\xmark}     &\red{\cmark}    &\red{\cmark}    &\red{\cmark}     &\green{\xmark} \\
            AdaptCLIP~\cite{adaptclip}          &\red{\cmark}   &\red{\cmark}  &\red{\cmark}   &\red{\cmark}     &\red{\cmark}    &\red{\cmark}    &\red{\cmark}     &\red{\cmark} \\
            
            \midrule
            \textbf{\method} (ours)            &\red{\cmark}  &\green{\xmark}  &\green{\xmark}  &\green{\xmark}     &\green{\xmark}    &\red{\cmark}    &\red{\cmark}     &\red{\cmark} \\
			 \bottomrule
		\end{tabular}
	}
\end{table}

A typical VLM is composed of a visual encoder $\mathcal{F}(\vec \theta_v;\cdot)$ and a text encoder $\mathcal{T}(\vec {\theta}_{t}; \cdot)$. For simplicity, we omit the original parameters, and $\vec \theta_v$ and $\vec \theta_t$ are optional, learnable parameters introduced for task adaptation in visual anomaly detection.
Given a query image $X_q\in \mathcal{R}^{H \times W \times 3}$, the visual encoder extracts its representations, yielding a global image token $\vec x_q$ and local patch tokens $F_q$:
\begin{equation}\label{eq:visionencoder}
    \vec x_q, F_q = \mathcal{F}(\vec \theta_v; X_q),
\end{equation}
where $\vec x_q \in \mathcal{R}^d$, $F_q \in \mathcal{R}^{d \times {H/p} \times {W/p}}$, $d$ is token dimension and $p$ is patch size.

The text encoder then generates a two-class weight matrix, $W \in \mathcal{R}^{d \times 2}$ using two-class textual prompts $T_f$ or learnable embeddings $T_l$, which embeds the semantics of ``normal'' and ``anomaly" concepts. 
\begin{equation}\label{eq:textencoder}
W = \mathcal{T}(\vec \theta_t; T_i),
\end{equation}
where $i=\{f,l\}$.
The pixel-level anomaly map, $\hat{Y}$, is derived by computing the similarity between $W$ and patch tokens $F^q$:
\begin{equation}\label{eq:patchsoftmax}
\hat Y_z = \softmax\big(\langle W, F^q \rangle / \tau\big).
\end{equation}
Here, $\langle \cdot \rangle$ denotes cosine similarity and $\tau > 0$ is a temperature parameter. The image-level anomaly score $\hat{y}$ is obtained by applying Eq.~\ref{eq:patchsoftmax} to the global image token $\vec x^q$:
\begin{equation}\label{eq:imagesoftmax}
\hat y_z = \softmax \big(\langle W, \vec x^q \rangle / \tau\big).
\end{equation}

As summarized in Tab.~\ref{tab:rethinksota}, recent methods are heavily dependent on text encoder, which limits the scalability, making them applicable only to VLMs and unable to be extended to other underlying models, such as DINOv3.
Furthermore, these existing approaches tend to aggregate multiple components to achieve competitive performance, which often leads to increasing model complexity. In contrast, we want to explore the feasibility of achieving a simple, general, and scalable AD framework by significantly reducing these components.

We can see that anomaly classification~(Eq.~\ref{eq:imagesoftmax}) and segmentation~(Eq.~\ref{eq:patchsoftmax}) are performed by a shared weight $W$. The $W$ currently is obtained either by feeding manual text prompts or learnable embeddings into text encoder~(Eq.~\ref{eq:textencoder}). This structure suggests that it is possible to obtain $W$ directly by learning it, thereby removing the text encoder and all associated textual components entirely. Thus, we have
\begin{equation}\label{eq:simpletextencoder}
W \leftarrow T_l.
\end{equation}
Modern foundation models possess strong generalization, which has been thoroughly demonstrated in many downstream visual tasks. Therefore, the learnable vision parameters $\vec \theta_v$ may be redundant, and Eq.~\ref{eq:visionencoder} is simplified as:
\begin{equation}\label{eq:simplevisionencoder}
    \vec x_q, F_q = \mathcal{F}(X_q).
\end{equation}
Substituting Eqs.~\ref{eq:simpletextencoder} and~\ref{eq:simplevisionencoder} into Eqs.~\ref{eq:patchsoftmax} and~\ref{eq:imagesoftmax}, we implement a simple and general zero-shot AD framework that only needs to learn weights $W$ based on features extracted from foundation models by removing most components as shown in Tab.~\ref{tab:rethinksota}.

\begin{figure}[thbp]
    \centering
    \begin{subfigure}[b]{0.23\textwidth}
        \centering
        \includegraphics[width=\linewidth, keepaspectratio]{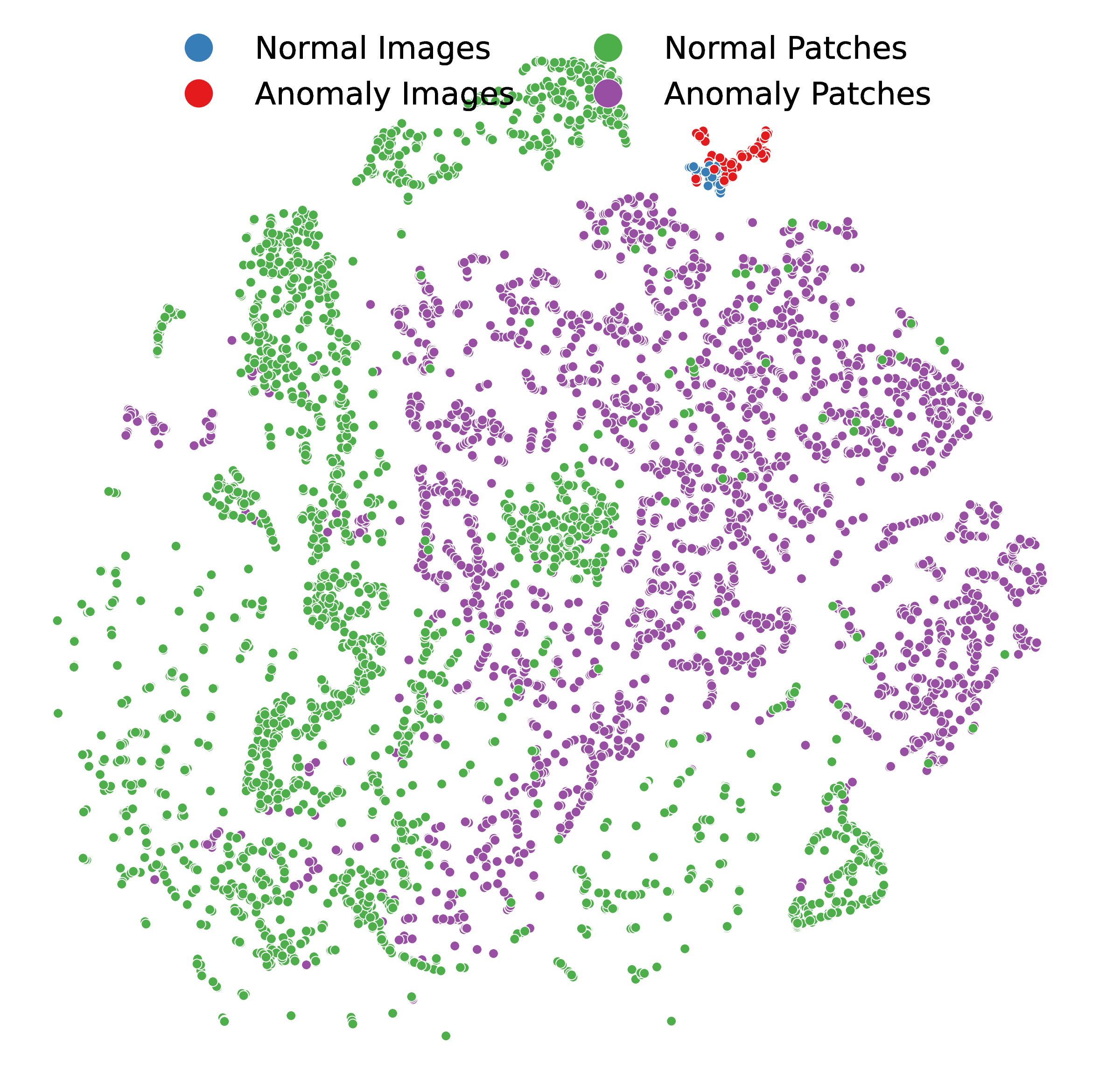}
        \caption{}
        \label{fig:manifold-a}
    \end{subfigure}
    \hfill
    \begin{subfigure}[b]{0.23\textwidth}
        \centering
        \includegraphics[width=\linewidth, keepaspectratio]{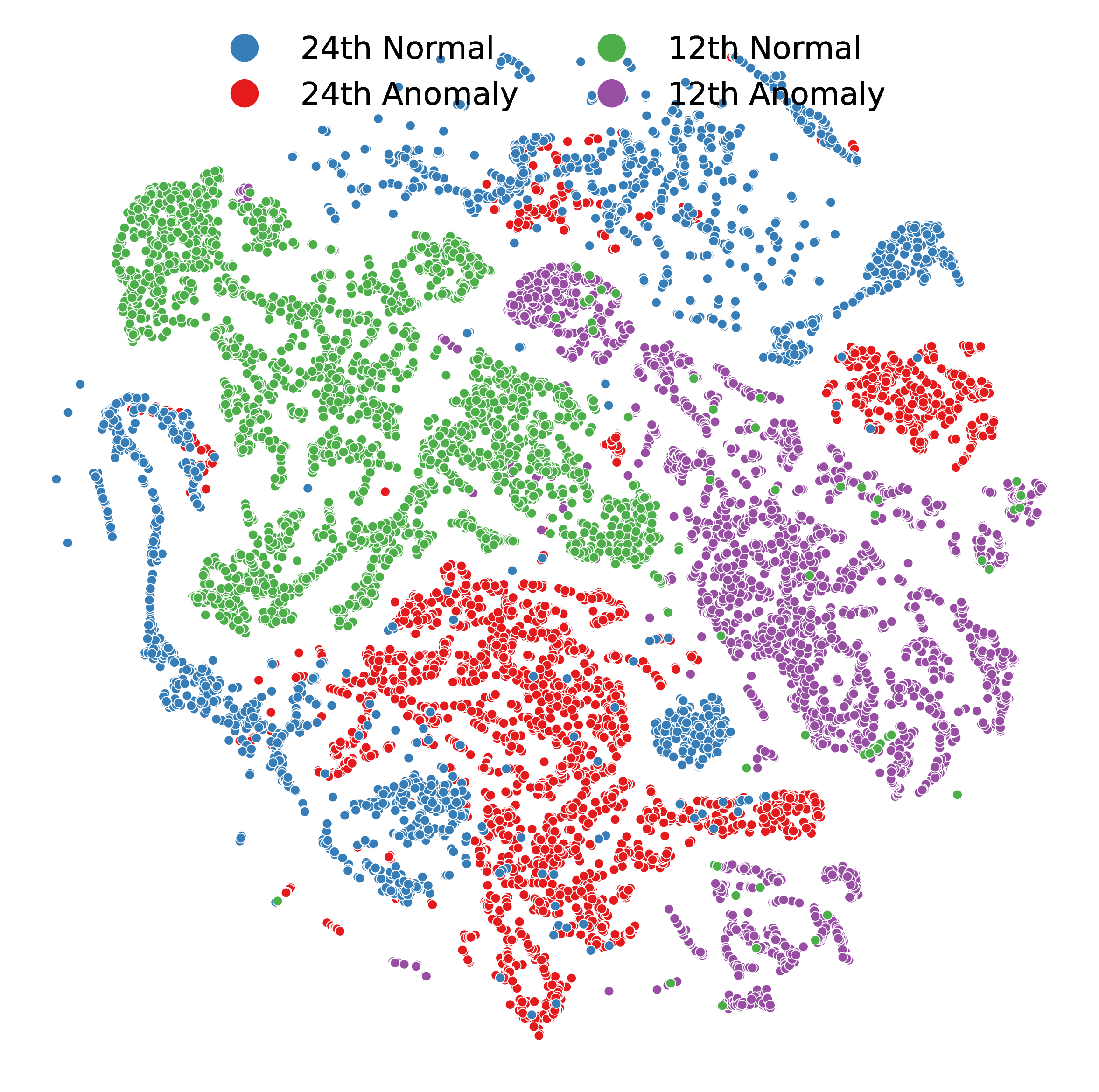}
        \caption{}
        \label{fig:manifold-b}
    \end{subfigure}
    \vspace{-8pt}
    \caption{\small{\small{t-SNE visualization of  CLIP ViT-L/14@336px features on MVTec Test Set (Hazelnuts). 
    (a) t-SNE embeddings of global image tokens ($\vec x^q$) and local patch tokens ($F^q$) extracted from the $24$-th block. The visualization clearly shows a significant disparity between the normal/anomaly distributions of image (global) tokens and patch (local) tokens. 
    (b) t-SNE embeddings of local patch tokens extracted from the $6$-th and $24$-th blocks. The normal and anomaly feature distributions across these two distinct hierarchical layers are also substantially different.}}}
    \label{fig:manifold}
\end{figure}

\subsection{Decoupling Classification and Segmentation}
\label{subsec:dcs}
The image anomaly classification and pixel anomaly segmentation commonly share the same weight $W$, \ie, Eqs.~\ref{eq:patchsoftmax} and~\ref{eq:imagesoftmax}, in several state-of-the-art zero-shot ADs, such as~\cite{iclr2024anomalyclip,eccv24adaclip,bayes-pfl,adaptclip}. However, it may be suboptimal because the manifolds of global image tokens and local patch tokens are fundamentally different. As shown in Fig.~\ref{fig:manifold}(a), the visualization empirically reveals that the decision hyperplane for normal and anomaly tokens differs significantly between these two representations (images and patches).  
Crucially, this divergence is observed consistently across both vision-language models (\eg, CLIP) and self-supervised purely vision models (\eg, DINOv2 and DINOv3).

To explicitly address the learning conflict between image anomaly classification and pixel anomaly segmentation, we decouple the single shared weight $W$ into a pair of specialized weights, $W_{cls}$ and $W_{seg}$. Consequently, Eqs.~\ref{eq:patchsoftmax} and \ref{eq:imagesoftmax} are transformed into a decoupled formulation:
\begin{equation}\label{eq:desoftmax}
\begin{aligned}
\hat Y_z &= \softmax\big(\langle W_{seg}, F_q \rangle / \tau\big),\\
\hat y_z &= \softmax\big(\langle W_{cls}, \vec x_q \rangle / \tau\big).
\end{aligned}
\end{equation}
We highlight two significant advantages. First, the decoupled weights fundamentally resolve the learning conflict caused by different manifolds between global and local tokens. 
Second, the proposed decoupling classification and segmentation is simple and efficient than existing SOTA methods using complex vision or textual adapters.

\subsection{Decoupling Hierarchical Features}\label{subsec:dhf}
For dense visual perception tasks, such as object detection and semantic segmentation, it is common practice to leverage multi-scale features extracted from a foundation model. Most zero-/few-shot AD works also use multi-scale features and share the same weight for these hierarchical features.

Different-scale features naturally encode varying semantics and thus embed in different manifolds as shown in Fig.~\ref{fig:manifold}(b). 
To effectively mitigate the conflict arising from differences of hierarchical features, we extend our decoupling concept to all features from different stages ot blocks. This ultimately results in multiple-layer-specific decoupling. 
Let $\vec x^l_q$ and $F^l_q$ denote global image tokens and local patch tokens extracted from the $l$-th layer, and $W^l_{seg}$ and $W^l_{cls}$ correspond to layer-specific weights. Consequently, the decoupled formulation ~(Eq.~\ref{eq:desoftmax}) at the $l$-th layer is further refined into the following form:
\begin{equation}\label{eq:fulldesoftmax}
\begin{aligned}
\hat Y^l_z &= \softmax\big(\langle W^l_{seg}, F^l_q \rangle / \tau\big), \\\
\hat y^l_z &= \softmax \big(\langle W^l_{cls}, \vec x^l_q \rangle / \tau\big).
\end{aligned}
\end{equation}
This complete decoupling ensures that anomaly classification and segmentation at each layer learns their optimal decision boundary without interference from disparate feature manifolds.

\subsection{\method~with Few-Shot Normal Images}
The \method~framework, utilizing the decoupled weights $\{W^l_{seg}\}$ and $\{W^l_{cls}\}$, is effective in detecting general, common anomalies (\eg, scratches or holes), but its capability is limited for some anomalies that are only defined relative to specific normal reference images (\eg, a missing battery cell).

Inspired by the memory bank in PatchCore~\cite{cvpr2022patchcore} and the reference association in WinCLIP~\cite{cvpr2023winclip}, we build a multi-scale normal memory to extend our framework to a few-shot setting.
We first store the local patch features at the $l$-th scale for $K$ ($\{1, 2, 4\}$) normal images, $X_p \in \mathcal{R}^{K \times H \times W \times 3}$, into a memory bank $M^l=\{F^l_p | F^l_p \in \mathcal{R}^{K\frac{HW}{P^2} \times d}\}$.

For a query image $X_q$, we extract its corresponding scale feature $F^l_q \in \mathcal{R}^{\frac{H}{P} \times \frac{W}{P} \times d}$. 
The few-shot anomaly score $\hat Y^l_f (i,j)$ is computed by cosine distance between the query feature and its nearest neighbor in $M^l$:
\begin{equation}\label{eq:fsmemory}
    \hat Y^l_f (i,j) = \min_{m^l_k \in M^l} \big(1 - \langle  {F^l_q (i,j)}, {m^l_k} \rangle \big),
\end{equation}
where $(i,j)$ is a spatial location.  
This process is performed on all scales (denoted as $L$) features. We then aggregate the multi-scale few-shot predictions using a simple average to obtain the final few-shot anomaly map:
\begin{equation}\label{eq:fsmap}
    \hat Y_{f} = \frac{1}{|L|} \sum_{l \in L} \hat Y^{l}_f. 
\end{equation}
Finally, we update $\hat Y_{f}$ by fusing the zero-shot prediction $\hat Y_{z}$ with our decoupled weights:
\begin{equation}\label{eq:fsmapfinal}
\hat Y_f \leftarrow (1-{\lambda}_{f})\hat Y_{z} + {\lambda}_{f} \hat Y_{f}, 
\end{equation}
where ${\lambda}_{f}>0$ is a balance factor that controls the trade-off between zero-shot and few-shot predictions. The few-shot image score is obtained by fusing the zero-shot score and the maximum value of the few-shot anomaly map, as 
\begin{equation}\label{eq:fsscorefinal}
\hat y_f \leftarrow (1-{\lambda}_{p})\hat y_{z} + {\lambda}_{p}  \max_{i,j}(\hat Y_f (i,j)).
\end{equation}

\subsection{Training and Inference}
Following AnomalyCLIP~\cite{iclr2024anomalyclip}, we optimize $\{W^l_{cls}\}$ using cross-entropy loss for global image anomaly classification, and $\{W^l_{seg}\}$ using Focal and Dice losses for local patch anomaly segmentation.  
In addition, multi-scale context plays a crucial role in visual anomaly detection because anomalies are essentially defined by their inconsistency with the context. We introduce a Class-Aware image Augmentation (CAA) to improve robustness (details in \texttt{Appendix}).

For few-shot inference, we use Eqs~\ref{eq:fsmapfinal} and~\ref{eq:fsscorefinal} to obtain pixel-level anomaly maps and image-level anomaly scores, respectively. For zero-shot inference, we use Eq~\ref{eq:fulldesoftmax} to compute pixel-level anomaly segmentation. Similar to Eq.~\ref{eq:fsscorefinal}, we also fuse the zero-shot score, \ie, $\hat y^l_z$, and the maximum value of the zero-shot anomaly map, \ie, $\max_{i,j} (\hat Y^l_z)$, with a weight factor $\lambda_p$. We set $\lambda_p$ and $\lambda_f$ to 0.5 as default.

%% file: sec/4_exps.tex
\begin{table*}[t]
	\caption{Comparisons with state-of-the-art zero-shot AD methods on industrial domain. \red{\cmark} indicates language-free and \green{\xmark} means not language-free, the best mean results are marked in \red{red}, while the second-best are indicated in \blue{blue}, and $^\dagger$, $^\ddagger$ and $^\star$ indicate CLIP ViT-L/14@336px, DINOv2-R ViT-L/16 and DINOv3 ViT-L/16, respectively, the same below.}
    \vspace{-8pt}
	\centering
	\label{tab:sota_zs_industrial}
	\renewcommand{\arraystretch}{1.0}
	\resizebox{1.0\textwidth}{!}
	{
		\begin{tabular}
        {
        >{\centering\arraybackslash}m{1cm}
        >{\centering\arraybackslash}m{1.8cm} 
        *{5}{>{\centering\arraybackslash}p{1.9cm}}
        >{\columncolor{lightgreen}\centering\arraybackslash}m{1.8cm}
        >{\columncolor{lightgreen}\centering\arraybackslash}m{1.8cm}
        >{\columncolor{lightgreen}\centering\arraybackslash}m{1.8cm}
        }
		\toprule
	\multirow{2}{*}{Metric}   &\multirow{2}{*}{Dataset}    &WinCLIP\cite{cvpr2023winclip}    &AdaCLIP\cite{eccv24adaclip}          &\small{AnomalyCLIP}\scriptsize{\cite{iclr2024anomalyclip}}          &\small{Bayes-PFL}\scriptsize{\cite{bayes-pfl}}  &AdaptCLIP\scriptsize{\cite{adaptclip}}              & \textbf{\method}$^\dagger$      & \textbf{\method}$^\ddagger$ & \textbf{\method}$^\star$         \\ 
&    &\green{\xmark}  &\green{\xmark} &\green{\xmark} &\green{\xmark} &\green{\xmark} &\red{\cmark} &\red{\cmark} &\red{\cmark} \\
    \midrule
\multirow{7}{*}{\rotatebox[origin=c]{90}{%
    \makecell[c]{Image-Level \\ (AUROC, AUPR)}%
}}
&MVTec	&90.4	, 95.6	&90.7	, 95.6	&91.6	, 96.4	&92.3	, 96.7	&93.5	, 96.7	&92.4	, 96.1	&93.5	, 97.2	&94.0	, 97.1\\
&VisA	&75.5	, 78.7	&81.7	, 84.3	&82.0	, 85.3	&87.0	, 89.2	&84.8	, 87.6	&88.0	, 89.1	&91.3	, 91.0	&91.9	, 92.8\\
&BTAD	&68.2	, 70.9	&89.9	, 95.5	&88.3	, 88.2	&93.2	, 96.5	&91.0	, 92.2	&96.4	, 97.8	&94.4	, 96.3	&94.7	, 97.0\\
&DTD	&95.1	, 97.7	&92.7	, 96.4	&93.9	, 97.2	&95.1	, 98.4	&96.0	, 98.4	&98.1	, 99.1	&94.3	, 98.2	&97.1	, 99.0\\
&KSDD	&92.9	, 84.9	&96.6	, 89.3	&97.8	, 94.2	&92.6	, 65.5	&98.1	, 95.7	&96.3	, 83.4	&91.6	, 59.5	&91.0	, 64.5\\
&Real-IAD	&67.0	, 62.9	&73.3	, 70.2	&69.5	, 64.3	&70.0	, 65.8	&74.2	, 70.8	&78.6	, 76.7	&82.5	, 80.3	&81.2	, 79.3\\
\cmidrule{2-10}
&\textbf{Mean}	&81.5	, 81.8	&87.5	, 88.6	&87.2	, 87.6	&88.4	, 85.3	&89.6	, 90.2	&\sbest{91.6} , \best{90.4}	&91.3, 87.1	&\best{91.7} , \sbest{88.3}\\
\midrule
\multirow{7}{*}{\rotatebox[origin=c]{90}{%
    \makecell[c]{Pixel-Level \\ (AUROC, AUPR)}%
}}
&MVTec	&82.3	, 18.2	&88.3	, 39.1	&91.1	, 34.5	&91.8	, 48.3	&90.9	, 38.3	&91.8	, 42.8	&92.4	, 50.9	&93.2	, 52.7\\
&VisA	&73.2	, 5.4	&95.7	, 31.0	&95.5	, 21.3	&95.6	, 29.8	&95.7	, 26.1	&95.8	, 28.0	&95.9	, 32.7	&96.3	, 32.5\\
&BTAD	&72.7	, 12.9	&91.6	, 42.9	&94.2	, 45.5	&93.9	, 47.1	&93.8	, 41.8	&95.7	, 42.8	&97.7	, 60.6	&97.8	, 58.9\\
&DTD	&79.5	, ~~9.8	&98.3	, 75.2	&97.9	, 62.6	&97.8	, 69.9	&97.7	, 68.7	&98.4	, 64.3	&98.3	, 71.9	&99.1	, 75.9\\
&KSDD	&93.0	, ~~7.1	&97.6	, 48.2	&98.1	, 51.9	&97.7	, 18.7	&98.1	, 58.1	&98.1	, 31.8	&98.4	, 15.8	&99.0	, 22.6\\
&Real-IAD	&84.5	, ~~3.3	&96.1	, 30.5	&95.1	, 26.7	&96.5	, 27.6	&94.9	, 28.2	&96.6	, 33.6	&97.7	, 43.1	&97.5	, 41.6\\
\cmidrule{2-10}
&\textbf{Mean}	&80.9	, ~~9.5	&94.6	, 44.5	&95.3	, 40.4	&95.5	, 40.2	&95.2	, 43.5	&96.1	, 40.6	&\sbest{96.7}, \sbest{45.8}	&\best{97.2} , \best{47.4}\\
    \bottomrule
	\end{tabular}
	}
\vspace{-8pt}
\end{table*}

\section{Experiments}\label{sec:exps}
\subsection{Experimental Setup}
\noindent\textbf{Datasets:}
We comprehensively evaluate \method~across two distinct domains covering industrial inspection and medical diagnosis.
In industrial domain, we evaluate six benchmarks with zero-shot and few-shot: MVTec~\cite{cvpr2019mvtec}, VisA~\cite{eccv2022visa}, BTAD~\cite{isie2021btad}, DTD~\cite{wacv2023dtd}, KSDD~\cite{jim2020ksdd}, and large-scale Real-IAD~\cite{cvpr2024realiad}. 
In medical domain, we only evaluate zero-shot on brain tumor detection (HeadCTs, BrainMRI~\cite{brainmri}, Br35H~\cite{Br35h}), skin lesion segmentation (ISIC~\cite{isic}), gastrointestinal polyp segmentation (ClinicDB~\cite{clinicdb}, ColonDB~\cite{colond}, Kvasir~\cite{icmm2020kvasir}, Endo~\cite{jhe2017endo}). Detailed statistical information for these datasets can be found in \texttt{Appendix}.

\noindent\textbf{Evaluation Metrics:} 
Following previous works, we use AUROC and AUPR for image anomaly classification and pixel anomaly segmentation in our main paper. Here, we emphasize that pixel AUPR is more suitable for anomaly segmentation, where the imbalance issue is very
extreme between normal and anomaly pixels~\cite{eccv2022visa}. In \texttt{Appendix}, we also provide detailed results using all metrics, including image and pixel AUROC, F1$_\text{max}$, AUPR, and pixel AUPRO.

\noindent\textbf{Training and Testing Protocol:} 
Following recent works~\cite{eccv24vcpclip,bayes-pfl}, we train \method~using the testing data from VisA and evaluate zero-/few-shot performance on other datasets. For the VisA evaluation, we use MVTec's test data to train \method. 

\noindent\textbf{Competing Methods:} We compare our \method~with diverse state-of-the-art zero-/few-shot AD methods including zero-shot WinCLIP~\cite{cvpr2023winclip}, AnomalyCLIP~\cite{iclr2024anomalyclip}, AdaCLIP~\cite{eccv24adaclip}, Baye-PFL~\cite{bayes-pfl} and few-shot WinCLIP+~\cite{cvpr2023winclip}, 
InCtrl~\cite{cvpr2024inctrl} and
AnomalyCLIP+. Here, AnomalyCLIP+ is a baseline adding patch-level feature associations to AnomalyCLIP~\cite{iclr2024anomalyclip}. More implementation details about \method~and competing methods can be found in \texttt{Appendix}.

\subsection{Comparisons with Zero-Shot Methods}
Tabs.~\ref{tab:sota_zs_industrial} and~\ref{tab:sota_zs_medical} compare \method~with SOTA zero-shot methods on two diverse domains, industrial and medical.

\begin{table*}[t]
	\caption{Comparisons with state-of-the-art zero-shot AD methods on medical domain.}
	\centering
	\label{tab:sota_zs_medical}
    \vspace{-8pt}
	\renewcommand{\arraystretch}{1.0}
	\resizebox{1.0\textwidth}{!}
	{
		\begin{tabular}
        {
        >{\centering\arraybackslash}m{1cm}
        >{\centering\arraybackslash}m{1.8cm} 
        *{5}{>{\centering\arraybackslash}p{1.9cm}}
        >{\columncolor{lightgreen}\centering\arraybackslash}m{1.8cm}
        >{\columncolor{lightgreen}\centering\arraybackslash}m{1.8cm}
        >{\columncolor{lightgreen}\centering\arraybackslash}m{1.8cm}
        }
		\toprule
	\multirow{2}{*}{Metric}   &\multirow{2}{*}{Dataset}    &WinCLIP\cite{cvpr2023winclip}    &AdaCLIP\cite{eccv24adaclip}          &\small{AnomalyCLIP}\scriptsize{\cite{iclr2024anomalyclip}}          &\small{Bayes-PFL}\scriptsize{\cite{bayes-pfl}}  &AdaptCLIP\scriptsize{\cite{adaptclip}}              & \textbf{\method}$^\dagger$      & \textbf{\method}$^\ddagger$ & \textbf{\method}$^\star$          \\ 
&    &\green{\xmark}  &\green{\xmark} &\green{\xmark} &\green{\xmark} &\green{\xmark} &\red{\cmark} &\red{\cmark} &\red{\cmark} \\
    \midrule
\multirow{4}{*}{\rotatebox[origin=c]{90}{%
    \makecell[c]{Image-Level \\ \tiny{(AUROC, AUPR)}}%
}}
&HeadCT	&83.7	, 81.6	&93.4	, 92.2	&93.0	, 92.6	&{96.5}	, {95.5}	&92.6	, 91.9	&{97.0}	, {96.7}	&95.2	, 94.9	&95.7	, 94.8\\
&BrainMRI	&92.0	, 90.7	&94.9	, 94.2	&90.2	, 92.4	&{96.2}	, 92.4	&90.1	, 92.0	&{95.7}	, {96.4}	&91.7	, 94.8	&91.9	, 95.0\\
&Br35H	&80.5	, 82.2	&95.7	, 95.7	&94.2	, 94.2	&{97.8}	, {96.2}	&94.8	, 95.1	&{97.4}	, {97.1}	&92.5	, 93.9	&92.3	, 93.8\\
\cmidrule{2-10}
&\textbf{Mean}	&85.4	, 84.8	&94.7	, 94.0	&92.5	, 93.1	&\best{96.8}	, \sbest{94.7}	&92.5	, 93.0	&\sbest{96.7}	, \best{96.7}	&93.1	, 94.5	&93.3	, 94.5\\
\midrule
\multirow{7}{*}{\rotatebox[origin=c]{90}{%
    \makecell[c]{Pixel-Level \\ \tiny{(AUROC, AUPR)}}%
}}
&ISIC	&83.3	, 62.4	&85.4	, 70.6	&89.4	, 76.2	&92.2	, 84.6	&91.0	, 80.9	&92.6	, 84.2	&95.0	, 88.5	&{95.9}	, {90.6}\\
&ColonDB	&64.8	, 14.3	&79.3	, 26.2	&81.9	, 31.7	&82.1	, 31.9	&{86.5}	, {52.0}	&84.0	, {36.2}	&86.0	, 35.5	&{86.7}	, 33.3\\
&ClinicDB	&70.7	, 19.4	&84.3	, 36.0	&82.9	, 38.2	&89.6	, 53.2	&83.8	, 40.2	&90.1	, 55.8	&{91.3}	, {59.3}	&{91.5}	, {57.5}\\
&Endo	&68.2	, 23.8	&84.0	, 44.8	&84.2	, 46.6	&89.2	, 58.6	&82.1	, 45.3	&90.7	, 62.5	&{92.6}	, {67.3}	&{92.8}	, {68.5}\\
&Kvasir	&69.8	, 27.5	&79.4	, 43.8	&79.1	, 39.8	&85.4	, 54.2	&86.5	, 52.0	&87.5	, 57.5	&{91.1}	, {65.2}	&{91.7}	, {65.9}\\
\cmidrule{2-10}
&\textbf{Mean}	&71.4	, 29.5	&82.5	, 44.3	&83.5	, 46.5	&87.7	, 56.5	&86.0	, 54.1	&89.0	, {59.2}	&\sbest{91.2}	, \best{63.2}	&\best{91.7}	, \best{63.2}\\
    \bottomrule
	\end{tabular}
	}
\vspace{-8pt}
\end{table*}

\noindent\textbf{Zero-Shot Generalization on Industrial Domain:}
As reported in Tab.~\ref{tab:sota_zs_industrial}, zero-shot results in industrial domain demonstrate the effectiveness and advantages of our language-free framework, highlighting three significant observations:
First, our language-free \method~substantially outperforms SOTA AdaptCLIP~\cite{adaptclip} ($91.6\%$ vs. $89.6\%$ I-AUROC) that relies on language prompts and text encoder in image anomaly classification, when using the same foundation model (\eg, CLIP). This validates our core hypothesis: complex language prompts, text encoders, and elaborate adaptation modules are unnecessary for achieving state-of-the-art ZS AD.
Second, the language-free nature of our framework enables a seamless, cost-free switch to highly capable vision foundation models, such as DINOv2 and DINOv3. When using a single DINOv2 or DINOv3, our method achieves absolute leading performance in pixel-level anomaly segmentation, surpassing SOTA Bayes-PFL~\cite{bayes-pfl} by a significant margin (\eg, $7\%$ P-AUPR, from $40.2\%$ to $47.4\%$). This empirically demonstrates that self-supervised vision foundation models are superior to language-supervised CLIP in extracting fine-grained representations, which are crucial for precise segmentation tasks.
Third, our \method~demonstrates exceptional robustness on the challenging, large-scale industrial AD benchmark, Real-IAD~\cite{cvpr2024realiad}. We achieve $82.5\%$ I-AUROC and $43.1\%$ P-AUPR, outperforming SOTA methods by about $8\%$ and $13\%$, respectively. This performance gain further substantiates the stability and scalability of our framework when tackling real-world and highly diverse datasets.

\noindent\textbf{Zero-Shot Generalization on Cross-Domain Medical}:
To further validate the generalization capability of our framework, we directly apply it, trained exclusively on an industrial auxiliary dataset (\eg, VisA), to medical anomaly detection (\ie, cross-domain generalization). As reported in Tab.~\ref{tab:sota_zs_medical}, our \method~consistently outperforms most SOTA approaches in eight medical benchmarks. Especially on pixel-level anomaly segmentation, our \method~with DINOv3 foundation model achieves $91.7\%$ P-AUROC and $63.2\%$ P-AUPR, substantially exceeding SOTA Bayes-PFL~\cite{bayes-pfl} by a large margin. This result decisively demonstrates the robust and superior cross-domain generalization of our language-free framework.

\begin{table}[t]
    \centering
    \setlength\tabcolsep{2.0pt}
    \caption{Complexity and efficiency comparisons.} \label{tab:sota_efficiency}
    \vspace{-8pt}
    \resizebox{0.48\textwidth}{!}{ 
        \begin{tabular}{@{}c ccccc@{}}
            \toprule
Shots	&Methods	&Models	&Input Size	&\# Params (M)	&Inf.Time (ms) \\
\midrule
\multirow{6}{*}{\textbf{0}}	
&AdaCLIP~\cite{eccv24adaclip}	&CLIP ViT-L/14@336px	&518$\times$518	&428.8 + 1.1e{\footnotesize{+}}1	&107.4 \\	
&AnomalyCLIP~\cite{iclr2024anomalyclip}	&CLIP ViT-L/14@336px	 &518$\times$518 	&427.9 + 5.6e{\footnotesize{+}}0	&~~70.7 \\
&Bayes-PFL~\cite{bayes-pfl} &CLIP ViT-L/14@336px	 &518$\times$518 	&427.9 + 2.7e{\footnotesize{+}}1	&154.9 \\
&AdaptCLIP-Zero	&CLIP ViT-L/14@336px	&518$\times$518	&427.9 + 6.0e-1	&~~57.5 \\
&\textbf{\method}$\dagger$ &CLIP ViT-L/14@336px	&518$\times$518	& 342.9 + \best{1.5e-3}	&\best{~~15.7} \\ 
&\textbf{\method}$^\star$  &DINOv3 ViT-L/16	&512$\times$512	& 303.2 + 2.0e-3	&~~41.9\\ 
\midrule
\multirow{5}{*}{\textbf{1}}
&InCtrl~\cite{cvpr2024inctrl}	&CLIP ViT-B-16+240	&240$\times$240 		&208.4 + 3.0e-1 &~~59.0 \\
&AnomalyCLIP+~\cite{iclr2024anomalyclip}	&CLIP ViT-L/14@336px	&518$\times$518	&427.9 + 5.6e{\footnotesize{+}}0	&~~76.2 \\
&AdaptCLIP &CLIP ViT-L/14@336px	&518$\times$518	&342.9 + 1.8e{\footnotesize{+}}0	&~~58.7\\ 
&\textbf{\method}$\dagger$ &CLIP ViT-L/14@336px	&518$\times$518	& 342.9 + \best{1.5e-3}	&\best{~~22.4}\\ 
&\textbf{\method}$^\star$ &DINOv3 ViT-L/16	&512$\times$512	& 303.2 + 2.0e-3	&~~48.4\\ 

             \bottomrule
        \end{tabular}
    } 
    \vspace{-8pt}
\end{table}

\begin{table*}[ht]
	\caption{Comparisons with state-of-the-art one-shot AD methods on industrial domain.}
	\centering
	\label{tab:sota_fs_industial}
    \vspace{-8pt}
	\renewcommand{\arraystretch}{1.0}
	\resizebox{1.0\textwidth}{!}
	{
		\begin{tabular}
        {
        >{\centering\arraybackslash}m{1.0cm}
        >{\centering\arraybackslash}m{2.3cm} 
        *{4}{>{\centering\arraybackslash}p{2.3cm}}
        >{\columncolor{lightgreen}\centering\arraybackslash}m{2.3cm}
        >{\columncolor{lightgreen}\centering\arraybackslash}m{2.3cm}
        >{\columncolor{lightgreen}\centering\arraybackslash}m{2.3cm}
        }
		\toprule
	\multirow{2}{*}{Metric}   &\multirow{2}{*}{Dataset}    &WinCLIP+\scriptsize{\cite{cvpr2023winclip}}    &InCtrl\scriptsize{\cite{cvpr2024inctrl}} &AnomalyCLIP+\scriptsize{\cite{iclr2024anomalyclip}}           &AdaptCLIP\scriptsize{\cite{adaptclip}}                & \textbf{\method}$^\dagger$      & \textbf{\method}$^\ddagger$ & \textbf{\method}$^\star$           \\ 
    &    &\green{\xmark}  &\green{\xmark} &\green{\xmark} &\green{\xmark}  &\red{\cmark} &\red{\cmark} &\red{\cmark} \\
    \midrule
\multirow{7}{*}{\rotatebox[origin=c]{90}{%
    \makecell[c]{Image-Level \\ (AUROC, AUPR)}%
}}
&MVTec	&\pmerror{93.6}{0.4} , \pmerror{96.8}{0.2}	&\pmerror{91.3}{0.4} , \pmerror{95.2}{0.3}	&\pmerror{95.2}{0.2} , \pmerror{97.2}{0.1}	&\pmerror{94.5}{0.5} , \pmerror{97.5}{0.1}	&\pmerror{96.3}{0.7} , \pmerror{98.1}{0.3}	&\pmerror{97.6}{0.2} , \pmerror{98.6}{0.3}	&\pmerror{97.2}{0.1} , \pmerror{98.1}{0.3}\\
&VisA	&\pmerror{80.0}{2.4} , \pmerror{81.7}{1.5}	&\pmerror{83.2}{2.4} , \pmerror{84.1}{1.5}	&\pmerror{86.1}{0.7} , \pmerror{87.7}{1.1}	&\pmerror{90.5}{1.2} , \pmerror{92.3}{0.9}	&\pmerror{89.3}{1.9} , \pmerror{90.6}{1.5}	&\pmerror{95.2}{0.8} , \pmerror{95.6}{0.6}	&\pmerror{95.8}{0.8} , \pmerror{95.5}{0.5}\\
&BTAD	&\pmerror{84.4}{1.5} , \pmerror{80.5}{3.3}	&\pmerror{88.5}{0.4} , \pmerror{83.4}{9.3}	&\pmerror{88.5}{0.8} , \pmerror{74.2}{1.5}	&\pmerror{93.4}{0.0} , \pmerror{95.8}{0.9}	&\pmerror{96.6}{0.3} , \pmerror{97.4}{0.4}	&\pmerror{96.4}{0.3} , \pmerror{97.7}{0.4}	&\pmerror{97.1}{0.5} , \pmerror{98.5}{0.2}\\
&DTD	&\pmerror{97.9}{0.2} , \pmerror{99.0}{0.1}	&\pmerror{97.9}{0.3} , \pmerror{98.9}{0.2}	&\pmerror{98.0}{0.2} , \pmerror{99.2}{0.2}	&\pmerror{98.0}{0.0} , \pmerror{99.1}{0.0}	&\pmerror{98.0}{0.2} , \pmerror{99.1}{0.1}	&\pmerror{98.8}{0.1} , \pmerror{99.5}{0.0}	&\pmerror{99.4}{0.1} , \pmerror{99.6}{0.0}\\
&KSDD	&\pmerror{93.8}{0.4} , \pmerror{84.6}{0.9}	&\pmerror{92.0}{0.9} , \pmerror{81.5}{2.2}	&\pmerror{97.5}{0.3} , \pmerror{95.2}{0.1}	&\pmerror{96.9}{0.3} , \pmerror{91.8}{0.2}	&\pmerror{97.7}{0.7} , \pmerror{92.4}{3.2}	&\pmerror{94.7}{0.4} , \pmerror{68.7}{2.0}	&\pmerror{96.9}{0.3} , \pmerror{84.7}{2.0}\\
&Real-IAD	&\pmerror{74.7}{0.2} , \pmerror{71.2}{0.3}	&\pmerror{76.6}{0.0} , \pmerror{69.9}{0.0}	&\pmerror{78.2}{0.0} , \pmerror{76.7}{0.0}	&\pmerror{81.8}{0.3} , \pmerror{80.4}{0.2}	&\pmerror{82.5}{0.6} , \pmerror{80.9}{0.6}	&\pmerror{88.7}{0.5} , \pmerror{86.8}{0.3}	&\pmerror{86.9}{0.6} , \pmerror{85.3}{0.5}\\
\cmidrule{2-9}
&\textbf{Mean}	&87.4 , 85.6	&88.3 , 85.5	&90.6 , 88.4	&92.5 , 92.8	&93.4 , \sbest{93.1}	&\best{95.2} , 91.2	&\best{95.6} , \best{93.6}\\
\midrule
\multirow{7}{*}{\rotatebox[origin=c]{90}{%
    \makecell[c]{Pixel-Level \\ (AUROC, AUPR)}%
}}
&MVTec	&\pmerror{93.4}{0.2} , \pmerror{38.3}{0.8}	&\pmerror{94.6}{0.2} , \pmerror{47.8}{1.1}	&\pmerror{92.8}{0.0} , \pmerror{40.8}{0.1}	&\pmerror{94.3}{0.1} , \pmerror{53.7}{0.9}	&\pmerror{95.6}{0.3} , \pmerror{55.7}{1.3}	&\pmerror{97.1}{0.1} , \pmerror{63.1}{0.5}	&\pmerror{97.4}{0.1} , \pmerror{66.5}{0.9}\\
&VisA	&\pmerror{94.7}{0.1} , \pmerror{15.8}{0.2}	&\pmerror{89.0}{0.2} , \pmerror{17.7}{0.6}	&\pmerror{96.4}{0.1} , \pmerror{24.8}{0.9}	&\pmerror{96.8}{0.0} , \pmerror{38.9}{0.3}	&\pmerror{97.5}{0.0} , \pmerror{32.8}{0.4}	&\pmerror{98.2}{0.1} , \pmerror{42.1}{0.2}	&\pmerror{98.2}{0.1} , \pmerror{44.0}{0.1}\\
&BTAD	&\pmerror{95.6}{0.2} , \pmerror{41.3}{2.6}	&\pmerror{96.6}{0.1} , \pmerror{44.1}{1.4}	&\pmerror{95.3}{0.3} , \pmerror{41.3}{1.1}	&\pmerror{96.6}{0.2} , \pmerror{60.6}{1.0}	&\pmerror{96.2}{0.1} , \pmerror{54.4}{2.5}	&\pmerror{98.4}{0.0} , \pmerror{72.7}{0.7}	&\pmerror{98.3}{0.1} , \pmerror{71.7}{1.0}\\
&DTD	&\pmerror{96.5}{0.1} , \pmerror{47.8}{0.9}	&\pmerror{98.6}{0.1} , \pmerror{64.3}{0.5}	&\pmerror{97.6}{0.1} , \pmerror{67.4}{0.4}	&\pmerror{97.4}{0.0} , \pmerror{76.9}{0.1}	&\pmerror{97.4}{0.1} , \pmerror{73.4}{0.9}	&\pmerror{98.9}{0.1} , \pmerror{82.4}{0.3}	&\pmerror{99.1}{0.0} , \pmerror{85.8}{0.2}\\
&KSDD	&\pmerror{97.6}{0.1} , \pmerror{19.2}{0.3}	&\pmerror{97.8}{0.2} , \pmerror{26.7}{0.7}	&\pmerror{98.6}{0.1} , \pmerror{47.5}{0.5}	&\pmerror{98.2}{0.1} , \pmerror{57.8}{1.2}	&\pmerror{98.0}{0.0} , \pmerror{45.0}{5.8}	&\pmerror{99.4}{0.3} , \pmerror{34.9}{2.7}	&\pmerror{99.3}{0.1} , \pmerror{45.2}{3.7}\\
&Real-IAD	&\pmerror{95.0}{0.0} , \pmerror{13.9}{0.2}	&\pmerror{95.4}{0.0} , \pmerror{19.1}{0.0}	&\pmerror{96.5}{0.0} , \pmerror{27.9}{0.0}	&\pmerror{97.1}{0.0} , \pmerror{36.6}{0.1}	&\pmerror{98.0}{0.0} , \pmerror{35.3}{0.2}	&\pmerror{98.9}{0.0} , \pmerror{48.4}{0.4}	&\pmerror{98.6}{0.1} , \pmerror{47.6}{0.5}\\
\cmidrule{2-9}
&\textbf{Mean}	&95.5 , 29.4	&95.3 , 36.6	&96.2 , 41.6 &96.7 , \sbest{54.1}	&97.1 , 49.4	&\best{98.5} , \sbest{57.3}	&\best{98.5} , \best{60.1}\\
    \bottomrule
	\end{tabular}
	}
\vspace{-8pt}
\end{table*}
\begin{table}[ht]
    \centering
    \setlength\tabcolsep{5.0pt}
    \caption{Comparisons with state-of-the-art few-/full-shot AD methods. {\gray{Gray}} indicates models trained with full normal images.}
    \label{tab:sota_full_industrial}
    \vspace{-8pt}
    \resizebox{0.48\textwidth}{!}{
    \begin{tabular}{@{}lccccc@{}}
        \toprule
        Methods    &Shots  & MVTec         & VisA          & Real-IAD          \\ \midrule
        \multirow{3}{*}{\textbf{\method}$^\ddagger$} 
        &1  & 97.6 / 63.1   & 95.2 / 42.1   &88.7 / 48.4  \\
        &2  & 98.0 / 64.1   & 96.1 / 44.2 & 89.0 / 46.7    \\
        &4  &\best{98.7} / \best{65.4}   & \best{96.9} / \best{45.2} &\best{90.3} / \best{48.5}   \\
    \midrule
        UniVAD~\cite{gu2025univad} & 1 & 97.8 / 55.6   & 93.5 / 42.8 & 85.1 / 37.6 \\
        \multirow{3}{*}{AdaptCLIP\cite{adaptclip}} &1  & 94.5 / 53.7   & 90.5 / 38.9   &81.8 / 36.6   \\
       &2  & 95.7 / 55.1    &92.2 / 40.7 & 82.9 / 37.8    \\
         &4  &96.6 / 57.2   & 93.1 / 41.8   & 83.9  /  39.1    \\
         MetaUAS~\cite{metauas} & 1 &90.7 / 59.3 & 81.2 / 42.7 & - / - \\
    \midrule
    \gray{Dinomaly~\cite{dinomaly}} &\gray{full} &\gray{99.6 / 69.3}  & \gray{98.7 / 53.2}   & \gray{89.3 / 42.8}   \\
    \gray{UniAD~\cite{neurips2022uniad}} &\gray{full} &\gray{96.5 / 44.7}  & \gray{90.8 / 33.6}   &\gray{83.0 / 21.1}   \\
    
     \bottomrule
    \end{tabular}}
    \label{tab:commanyshots}
    \vspace{-8pt}
\end{table}

\noindent\textbf{Complexity and Efficiency:}
We measure complexity using the number of model parameters (frozen foundation model $+$ learnable components), and efficiency with
forward inference time of one image, as shown in Tab.~\ref{tab:sota_efficiency}. The evaluation is performed on one H20 GPU with batch size 1. The number of learnable parameters of our \method~is reduced by 10,000$\times$ and 300$\times$ compared to SOTA Bayes-PFL~\cite{bayes-pfl} and AdaptCLIP~\cite{adaptclip}, respectively.
Our \method~achieves the best inference speed, \eg, 15.7ms and 22.4ms with CLIP ViT-L/14@336px model. In addition, our \method~requires no additional parameters and almost no additional inference time when extending from zero-shot to one-shot.

\begin{figure}[htbp]
    \centering
    \includegraphics[width=1.0\columnwidth, keepaspectratio]{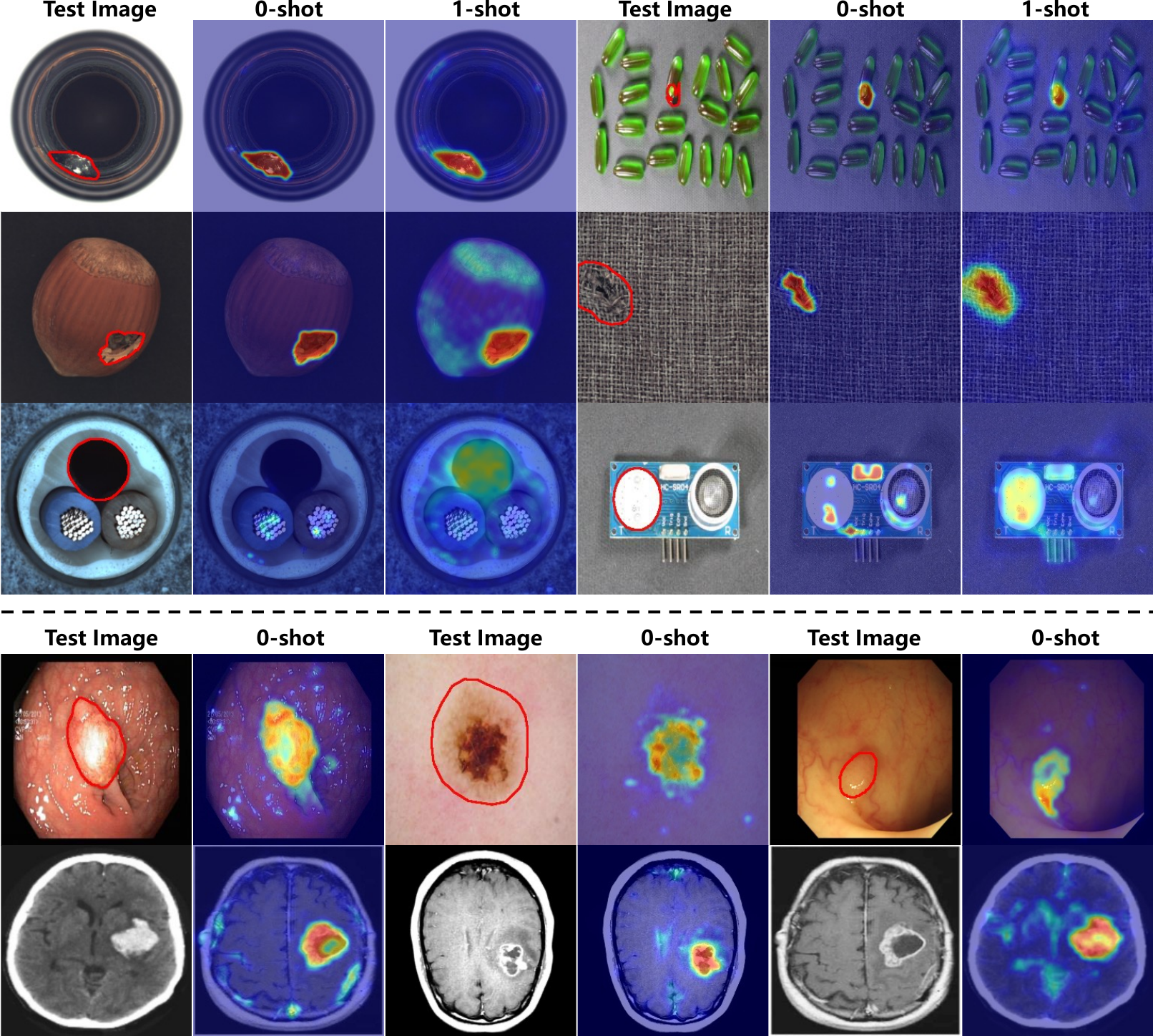} 
    \vspace{-10pt}
    \caption{\small{Qualitative comparisons of \method~using foundation DINOv3 (ViT-L/16) on industrial and medical domains. Best viewed in color and zoom.
    }}
    \label{fig:viszsfs}
    \vspace{-10pt}
\end{figure}

\noindent\textbf{Qualitative Results:}
Fig.~\ref{fig:viszsfs} shows some selected visualizations from industrial and medical testing images using \method~with a foundation DINOv3 (ViT-L/16). Generally, one-shot mode improves the performance of zero-shot, especially for defects that are defined by a normal image reference (shown in the third row).

\subsection{Comparisons with Few-/Full-Shot Methods}
In Tabs.~\ref{tab:sota_fs_industial} and~\ref{tab:sota_full_industrial}, we compare one- and few-shot \method~with few-shot and full-shot models.
It can be seen that there is a substantial performance gain across both image-level classification and pixel-level segmentation metrics for all methods. This indicates that one normal reference image is critical in industrial AD, as some defects are inherently defined by their deviation from normal images. Our results in the few-shot regime reveal several key advantages.

\emph{SOTA performance with minimal resources}: our 1-shot \method~achieves superior performance, such as $95.6\%$ I-AUROC and $98.5\%$ P-AUROC in Tab.~\ref{tab:sota_fs_industial}, by solely employing a single pure vision foundation model, \eg DINOv3, and training-free multi-scale normal memory bank. This significantly surpasses those methods that leverage VLMs, highlighting the efficiency of our decoupling and memory strategy.
\emph{Competitive with model ensembles}: our method with only a single foundation model (DINOv2) achieves comparable performance as well as recent ensemble UniVAD~\cite{gu2025univad} in Tab.~\ref{tab:sota_full_industrial}, which integrates multiple powerful models (RAM, Grounding DINO, SAM, CLIP, and DINOv2). This underscores the architectural and computational efficiency of our design.
\emph{Challenging SOTA full-shot methods}: with just a few training-free normal images (\eg, 4), our performance significantly surpasses earlier full-shot methods such as UniAD~\cite{neurips2022uniad}. Furthermore, our results are highly competitive with the current best fully-shot Dinomaly~\cite{dinomaly} ($98.7\%$ vs. $99.6\%$ in I-AUROC on MVTec). We exceed Dinomaly~\cite{dinomaly} on the challenging Real-IAD ($90.3\%$ vs. $89.3\%$ in I-AUROC).
It is noteworthy that our superior performance is achieved while maintaining a training-free and parameter-efficient setup. This clearly demonstrates the powerful potential of foundation models for few-shot anomaly detection in open-world scenarios.

\subsection{Ablation Study}

To demonstrate the effectiveness of our \method,  we perform ablation studies for core components: Decoupling Classification and Segmentation (DCS),  Decoupling Hierarchical Features (DHF), and Class-Aware Augmentation (CAA). We conduct experiments on MVTec and VisA, and report results in Tab.~\ref{tab:ab}. In addition, we analyze the effect on hierarchical features in Tab.~\ref{tab:multi-scale}.

\noindent\textbf{DCS}:
We first establish a simple baseline (Line $\text{\color{gray}0}$) by removing CLIP text encoder and utilizing a shared weight for both global anomaly classification and local anomaly segmentation across all feature levels. This baseline already outperforms AnomalyCLIP~\cite{iclr2024anomalyclip} in pixel-level. Crucially, when we decouple anomaly classification and segmentation learning two independent weights (Line $\text{\color{gray}1}$ vs. $\text{\color{gray}0}$), both image- and pixel-level performance improve significantly. This empirically verifies that the decoupling strategy effectively mitigates the inherent conflict between the two tasks, which arises from their distinct feature manifolds.

\noindent\textbf{DHF}:
We further extend the decoupling concept to encompass all feature levels, where every scale feature utilizes its own independent weight (Line $\text{\color{gray}2}$). Correspondingly, the performance is observed to improve yet again. This progressive gain conclusively confirms that the full decoupling mechanism effectively mitigates task conflicts arising from disparate feature manifolds both within a layer (global vs. local) and across different layers. Ultimately, this decoupling strategy successfully unlocks the full potential of the foundation model for universal anomaly detection.

\noindent\textbf{CAA}:
The class-aware augmentation leads to a further performance enhancement by multi-scale context (Line $\text{\color{gray}3}$). In contrast, we find that employing a class-agnostic augmentation strategy (\ie, random in Line $\text{\color{gray}3}$) results in performance degradation. This empirical evidence confirms that maintaining semantic consistency across augmented context is vital for robust learning in anomaly detection.

\noindent\textbf{Impact on Hierarchical Features}.
We analyze the influence of integrating hierarchical features from different blocks on overall zero-shot AD performance, as reported in Tab.~\ref{tab:multi-scale}. The performance generally improves as more features. This validates a fundamental assumption that hierarchical features provide rich and complementary information, which is crucial for comprehensive AD. Fine-grained pixel-level segmentation relies highly on the low-level details preserved in shallower blocks (\eg, 12 and 15). 

\begin{table}[t]
	\caption{Ablation studies about different components.}
	\centering
    \setlength\tabcolsep{5pt}
	\label{tab:ab}
    \vspace{-8pt}
	\renewcommand{\arraystretch}{1.0}
	\resizebox{1.0\columnwidth}{!}
	{
		\begin{tabular}{c cccc cc}
		\toprule
No &DCS    &DHF    &CAA &Shot    &{MVTec}                &{VisA}          \\ 
    \midrule
\gray{0}	&\green{\xmark}	&\green{\xmark}	&\green{\xmark}	&0	&85.4 / 36.4  	&77.9 / 26.1 \\
\gray{1}	&\red{\cmark}	&\green{\xmark}	&\green{\xmark}	&0	&91.8 / 38.3 	&85.9 / 27.2 \\
\gray{2}	&\red{\cmark}	&\red{\cmark}	&\green{\xmark}	&0	&92.2 / 40.7 	&86.0 / 27.6 \\
\gray{3}	&\red{\cmark}	&\red{\cmark}	&\red{\cmark}	&0	&92.4 / 42.8  	&88.0 / 28.0 \\
\gray{4}    &\red{\cmark}	&\red{\cmark}	& \scriptsize{random}   &0  &91.3  / 41.5  &   87.5 / 26.6  \\
\gray{5}	&\red{\cmark}	&\red{\cmark}	&\red{\cmark}	&1	&95.9 / 54.6	&91.3 / 32.5 \\
\bottomrule
	\end{tabular}
	}
    \vspace{-8pt}
\end{table}

\begin{table}[t]
	\caption{Impact on hierarchical features.}
	\centering
    \setlength\tabcolsep{5pt}
	\label{tab:multi-scale}
    \vspace{-8pt}
	\renewcommand{\arraystretch}{1.0}
	\resizebox{1.0\columnwidth}{!}
	{
		\begin{tabular}{@{}cc c cc@{}}
		\toprule
No &Blocks    &Shot      &{MVTec}                &{VisA}          \\ 
    \midrule

\gray{1} &\{24\}		&0		&89.9 / 35.4  	&85.3 / 23.1  \\
\gray{2} &\{21, 24\}		&0		&92.1 / 36.7	&87.3 / 24.2  \\
\gray{3} &\{18, 21, 24\}		&0		&92.0 / 38.9  	&87.7 / 26.1 \\ 
\gray{4} &\{15, 18, 21, 24\}		&0		&92.2 / 41.1 	&87.6 / 26.9 \\ 
\gray{5} &\{12, 15, 18, 21, 24\}	&0			&92.4 / 42.8 	&88.0 / 28.0 \\ 
\bottomrule
	\end{tabular}
	}
    \vspace{-8pt}
\end{table}

%% file: sec/5_cons.tex
\section{Conclusions}\label{sec:cons}
This paper fundamentally challenges and rethinks most existing language-dependent vision anomaly detection methods. 
Our core insights suggest that the primary function of a language encoder is to derive decision weights for anomaly classification and segmentation.
Therefore, we believe that the language encoder may be unnecessary, as these weights can be learned directly and efficiently through a data-driven process, thus negating the need for prompt engineering, complex adaptation, and training strategies, which recent works have been pursuing.
This breakthrough allows us to present \method, a remarkably \emph{simple, general, and language-free} framework for universal anomaly detection.
Observing learning conflicts arising from different feature manifolds among multi-level features in anomaly classification and segmentation using shared weights, this paper proposes an embarrassingly simple method to completely decouple classification and segmentation, and decouple cross-level features, \ie, learning independent weights for different tasks corresponding to features at different levels. 
The language-free approach is more general, allowing us to use diverse foundation models, both vision-language CLIP and pure vision DINOv2 and DINOv3.
Extensive experiments on diverse anomaly detection domains covering industrial and medical confirm that our \method~achieves SOTA performance in both zero-shot and few-shot settings. Notably, on a large-scale real-world industrial AD benchmark Real-IAD, our method, using just 4 normal images and requiring no training,  outperforms full-shot SOTA for the first time. This reveals the powerful potential of language-free framework in visual anomaly detection.

%% file: sec/X_suppl.tex
\clearpage
\setcounter{page}{1}
\maketitlesupplementary

\section{Class-Aware Augmentation}\label{sec:cma}
Multi-scale context plays a crucial role in visual anomaly detection because the essence of anomalies lies in their inconsistency with the context. We propose a Class-Aware Augmentation (CAA) method, including Grid Mosaic and Grid Cropping, to enhance the model's robustness.

\noindent\textbf{Grid Mosaic:} Given a candidate image and its corresponding category, we construct an augmented image by randomly sampling $n^2-1$ additional images of the same category from training set into an $n \times n$ grid.
To ensure the image-level ratio of normal/anomaly images remains unbiased, we apply distinct sampling strategies based on the current image's label (\ie, normal or anomaly). If the candidate image is normal, we randomly sample normal images of the same category as mosaic components. If the candidate image is an anomaly, we randomly sample images of the same category, which may include a mix of both normal and anomaly images. This conditional sampling approach is crucial for maintaining the ratio of normal and anomaly images and focusing the learning process on the true contextual discrepancies that define an anomaly.

\noindent\textbf{Grid Cropping:} To achieve diverse scale representations and effectively simulate resolution reduction from combining multiple images, we introduce a Grid Cropping Augmentation. Specifically, the current image is first randomly partitioned into an $n \times n$ grid. Subsequently, we randomly select one of these resulting image patches to serve as the augmented input. To ensure the ratio of normal and anomaly distribution is preserved during this process, we employ different strategies for normal and anomaly images. For a normal image, we select a patch at random. For an anomaly image, we strictly select only those patches that contain an anomaly region. This focused sampling strategy ensures that the model receives reliable supervisory signals for localization while effectively learning features across various scales and resolutions.

\begin{table*}[t]
\setlength\tabcolsep{3pt}
\centering
\small
\caption{\small{Key statistics of industrial and medical datasets with different attributes. \cmark~means satisfied and \xmark~means not satified.}}\label{tab:datasets}
\label{tab:sota_aupr}
\vspace{-5pt}
\resizebox{1.0\textwidth}{!}{
\begin{tabular}{c|ccc cc cc cc cc}
\toprule
\multirow{2}{*}{\textbf{Domain}} 	&\multirow{2}{*}{\textbf{Dataset}} 	&\multirow{2}{*}{\textbf{Modality}}	&\multirow{2}{*}{\textbf{Category}}	&\multirow{2}{*}{\textbf{\# Classes}} &\multirow{2}{*}{\textbf{Pose-Agnostic}}	
&\multicolumn{2}{c}{\textbf{Anotations}}		&\textbf{Train}	&\multicolumn{2}{c}{\textbf{Test}}	\\			
&&&&& &\textbf{Image} &\textbf{Pixel} &\textbf{\# Normal} &\textbf{\# Normal} &\textbf{\# Anomaly} \\
\toprule
\multirow{6}{*}{\textbf{Industrial}}
&MVTec~\cite{cvpr2019mvtec} 	&Photography	&Obj \& Texture	&15	&\xmark	&\cmark	&\cmark	&~~3,629	&~~~~467	&~~1,258\\
	&VisA~\cite{eccv2022visa} 	&Photography	&Obj	&12	&\xmark	&\cmark	&\cmark	&~~8,659	&~~~~962	&~~1,200\\
	&BTAD~\cite{isie2021btad} 	&Photography	&Obj \& Texture	&3	&\xmark	&\cmark	&\cmark	&~~1,799	&~~~~451	&~~~~290\\
	&DTD~\cite{wacv2023dtd} 	&Photography	&Texture	&12	&\xmark	&\cmark	&\cmark	&~~1,200	&~~~~357	&~~~~947\\
	&KSDD~\cite{jim2020ksdd} 	&Photography	&Texture	&1	&\xmark	&\cmark	&\cmark	&~~~~857	&~~~~286	&~~~~~54\\
	&Real-IAD~\cite{cvpr2024realiad} 	&Photography	&Obj	&30	&\xmark	&\cmark	&\cmark	&36,465	&63,256	&51,329\\
    \midrule

\multirow{8}{*}{\textbf{Medical}}

&HeadCT~\cite{salehi2021multiresolution}  &CT &Head &1 &\cmark	&\cmark	&\xmark	&~~~~~~0	&~~~100	&~~~100 \\
&BrainMRI~\cite{brainmri} &MRI &Brain &1  &\cmark	&\cmark	&\xmark	&~~~~~~0	&~~~155	&~~~~~~98 \\
&Br35H~\cite{Br35h} &MRI &Brain Tumor &1  &\cmark	&\cmark	&\xmark	&~~~~~~0	&~~~1,500	&~~~1,500 \\
&ISIC~\cite{isic} &Endoscopy &Skin Lesion &1  &\cmark	&\xmark	&\cmark	&~~~~~~0	&~~~~~~0	&~~~~379 \\
&ColonDB~\cite{colond} &Endoscopy &Colorectal Cancer &1  &\cmark	&\xmark	&\cmark	&~~~~~~0	&~~~~~~0	&~~~~380 \\
&ClinicDB~\cite{clinicdb} &Endoscopy &Colorectal Cancer &1  &\cmark	&\xmark	&\cmark	&~~~~~~0	&~~~~~~0	&~~~~612 \\
&Endo~\cite{jhe2017endo}	&Endoscopy	&Gastrointestinal Tract	&1	&\cmark	&\xmark	&\cmark	&~~~~~~0	&~~~~~~0	&~~~~200\\
&Kvasir~\cite{icmm2020kvasir}	&Endoscopy	&Gastrointestinal Tract	&1	&\cmark	&\xmark	&\cmark	&~~~~~~0	&~~~~~~0	&~~1,000\\
\bottomrule
\end{tabular}
}
\vspace{0pt}
\end{table*}

\begin{table*}[ht]
	\caption{Comparisons with state-of-the-art 2-shot AD methods on industrial domain. \red{\cmark} indicates language-free and \green{\xmark} means not language-free, the best mean results are marked in \red{red}, while the second-best are indicated in \blue{blue}, and $^\dagger$, $^\ddagger$ and $^\star$ indicate CLIP ViT-L/14@336px, DINOv2-R ViT-L/16 and DINOv3 ViT-L/16, respectively, the same below.}
	\centering
	\label{tab:2s-industial}
    \vspace{-8pt}
	\renewcommand{\arraystretch}{1.0}
	\resizebox{1.0\textwidth}{!}
	{
		\begin{tabular}
        {
        >{\centering\arraybackslash}m{1.0cm}
        >{\centering\arraybackslash}m{2.3cm} 
        *{4}{>{\centering\arraybackslash}p{2.3cm}}
        >{\columncolor{lightgreen}\centering\arraybackslash}m{2.3cm}
        >{\columncolor{lightgreen}\centering\arraybackslash}m{2.3cm}
        >{\columncolor{lightgreen}\centering\arraybackslash}m{2.3cm}
        }
		\toprule
	\multirow{2}{*}{Metric}   &\multirow{2}{*}{Dataset}    &WinCLIP+\scriptsize{\cite{cvpr2023winclip}}    &InCtrl\scriptsize{\cite{cvpr2024inctrl}} &AnomalyCLIP+\scriptsize{\cite{iclr2024anomalyclip}}           &AdaptCLIP\scriptsize{\cite{adaptclip}}                & \textbf{\method}$^\dagger$      & \textbf{\method}$^\ddagger$ & \textbf{\method}$^\star$           \\ 
    &    &\green{\xmark}  &\green{\xmark} &\green{\xmark} &\green{\xmark}  &\red{\cmark} &\red{\cmark} &\red{\cmark} \\
    \midrule
\multirow{7}{*}{\rotatebox[origin=c]{90}{%
    \makecell[c]{Image-Level \\ (AUROC, AUPR)}%
}}
&MVTec	&\pmerror{94.5}{1.0}	, \pmerror{97.3}{0.5}	&\pmerror{91.8}{0.9}	, \pmerror{95.5}{0.7}	&\pmerror{95.4}{0.1}	, \pmerror{97.3}{0.1}	&\pmerror{95.7}{0.6}	, \pmerror{97.9}{0.2}	&\pmerror{97.0}{0.6}	, \pmerror{98.3}{0.2}	&\pmerror{98.0}{0.3}	, \pmerror{98.9}{0.1}	&\pmerror{97.6}{0.3}	, \pmerror{98.6}{0.2}\\
&VisA	&\pmerror{82.7}{1.0}	, \pmerror{84.0}{0.7}	&\pmerror{86.3}{1.4}	, \pmerror{86.8}{1.7}	&\pmerror{87.8}{0.5}	, \pmerror{89.1}{0.7}	&\pmerror{92.2}{0.8}	, \pmerror{93.6}{0.6}	&\pmerror{92.3}{0.5}	, \pmerror{93.0}{0.4}	&\pmerror{96.1}{0.8}	, \pmerror{96.3}{0.8}	&\pmerror{96.8}{0.4}	, \pmerror{96.3}{0.4}\\
&BTAD	&\pmerror{85.8}{1.8}	, \pmerror{82.5}{3.2}	&\pmerror{86.2}{2.0}	, \pmerror{81.6}{8.0}	&\pmerror{89.2}{1.1}	, \pmerror{75.4}{1.5}	&\pmerror{93.4}{0.2}	, \pmerror{95.9}{0.1}	&\pmerror{96.5}{0.2}	, \pmerror{97.3}{0.4}	&\pmerror{96.8}{0.4}	, \pmerror{98.2}{0.4}	&\pmerror{97.3}{0.6}	, \pmerror{98.6}{0.2}\\
&DTD	&\pmerror{98.1}{0.2}	, \pmerror{99.1}{0.1}	&\pmerror{98.3}{0.2}	, \pmerror{99.1}{0.3}	&\pmerror{98.2}{0.1}	, \pmerror{99.3}{0.1}	&\pmerror{97.4}{0.0}	, \pmerror{99.2}{0.0}	&\pmerror{98.0}{0.1}	, \pmerror{99.1}{0.1}	&\pmerror{99.2}{0.1}	, \pmerror{99.6}{0.0}	&\pmerror{99.6}{0.1}	, \pmerror{99.7}{0.1}\\
&KSDD	&\pmerror{93.8}{0.2}	, \pmerror{84.5}{0.6}	&\pmerror{91.6}{0.9}	, \pmerror{81.0}{2.5}	&\pmerror{97.9}{0.2}	, \pmerror{95.6}{0.2}	&\pmerror{97.2}{0.0}	, \pmerror{92.4}{0.4}	&\pmerror{97.7}{0.6}	, \pmerror{93.3}{1.3}	&\pmerror{95.3}{0.2}	, \pmerror{71.2}{1.2}	&\pmerror{97.2}{0.4}	, \pmerror{87.5}{0.9}\\
&RealIAD	&\pmerror{76.1}{0.1}	, \pmerror{72.7}{0.1}	&\pmerror{78.5}{0.0}	, \pmerror{71.7}{0.0}	&\pmerror{78.3}{0.0}	, \pmerror{76.9}{0.0}	&\pmerror{82.9}{0.2}	, \pmerror{81.5}{0.1}	&\pmerror{83.6}{0.2}	, \pmerror{82.0}{0.3}	&\pmerror{89.0}{0.7}	, \pmerror{86.7}{1.4}	&\pmerror{87.2}{1.2}	, \pmerror{85.0}{2.2}\\
\cmidrule{2-9}
&\textbf{Mean}	&88.5	, 86.7	&88.8	, 86.0	&91.1	, 88.9	&93.1	, 93.4	&94.2	, \sbest{93.8}	&\sbest{95.7}	, {91.8}	&\best{96.0}	, \best{94.3}\\
\midrule
\multirow{7}{*}{\rotatebox[origin=c]{90}{%
    \makecell[c]{Pixel-Level \\ (AUROC, AUPR)}%
}}
&MVTec	&\pmerror{93.8}{0.1}	, \pmerror{39.5}{0.6}	&\pmerror{95.2}{0.2}	, \pmerror{49.2}{0.7}	&\pmerror{92.9}{0.1}	, \pmerror{41.5}{0.1}	&\pmerror{94.5}{0.0}	, \pmerror{55.1}{0.5}	&\pmerror{95.9}{0.1}	, \pmerror{57.0}{0.9}	&\pmerror{97.4}{0.1}	, \pmerror{64.1}{0.6}	&\pmerror{97.7}{0.1}	, \pmerror{67.5}{1.0}\\
&VisA	&\pmerror{95.1}{0.1}	, \pmerror{17.2}{0.8}	&\pmerror{89.8}{0.2}	, \pmerror{18.5}{0.2}	&\pmerror{96.7}{0.1}	, \pmerror{26.2}{0.7}	&\pmerror{97.1}{0.0}	, \pmerror{40.7}{0.6}	&\pmerror{97.8}{0.0}	, \pmerror{35.8}{1.0}	&\pmerror{98.4}{0.0}	, \pmerror{44.2}{0.3}	&\pmerror{98.3}{0.0}	, \pmerror{45.3}{0.4}\\
&BTAD	&\pmerror{95.7}{0.1}	, \pmerror{42.8}{1.3}	&\pmerror{96.7}{0.1}	, \pmerror{44.2}{0.8}	&\pmerror{95.5}{0.2}	, \pmerror{41.9}{0.6}	&\pmerror{96.7}{0.1}	, \pmerror{61.0}{0.6}	&\pmerror{96.4}{0.1}	, \pmerror{54.5}{2.1}	&\pmerror{98.4}{0.0}	, \pmerror{73.1}{0.5}	&\pmerror{98.4}{0.0}	, \pmerror{71.8}{1.3}\\
&DTD	&\pmerror{96.6}{0.1}	, \pmerror{48.2}{0.9}	&\pmerror{98.7}{0.1}	, \pmerror{64.4}{0.4}	&\pmerror{97.7}{0.1}	, \pmerror{68.1}{0.2}	&\pmerror{97.6}{0.0}	, \pmerror{77.4}{0.2}	&\pmerror{97.6}{0.0}	, \pmerror{74.3}{0.7}	&\pmerror{99.0}{0.1}	, \pmerror{83.0}{0.1}	&\pmerror{99.2}{0.1}	, \pmerror{86.1}{0.2}\\
&KSDD	&\pmerror{97.6}{0.1}	, \pmerror{19.0}{0.5}	&\pmerror{97.6}{0.6}	, \pmerror{26.4}{2.5}	&\pmerror{98.6}{0.1}	, \pmerror{47.6}{0.4}	&\pmerror{98.1}{0.1}	, \pmerror{57.5}{1.1}	&\pmerror{98.1}{0.2}	, \pmerror{46.1}{3.9}	&\pmerror{99.5}{0.1}	, \pmerror{36.9}{2.7}	&\pmerror{99.3}{0.2}	, \pmerror{45.0}{2.2}\\
&RealIAD	&\pmerror{95.3}{0.0}	, \pmerror{14.8}{0.1}	&\pmerror{96.0}{0.0}	, \pmerror{20.1}{0.0}	&\pmerror{96.6}{0.0}	, \pmerror{28.1}{0.0}	&\pmerror{97.3}{0.0}	, \pmerror{37.8}{0.1}	&\pmerror{98.2}{0.0}	, \pmerror{36.4}{0.3}	&\pmerror{99.1}{0.2}	, \pmerror{46.7}{2.3}	&\pmerror{98.9}{0.2}	, \pmerror{47.0}{2.9}\\
\cmidrule{2-9}
&\textbf{Mean}	&95.7	, 30.3	&95.7	, 37.1	&96.3	, 42.2	&96.9	, 54.9	&\sbest{97.3}	, 50.7	&\best{98.6}	, \sbest{58.0}	&\best{98.6}	, \best{60.5}\\
    \bottomrule
	\end{tabular}
	}
\vspace{-8pt}
\end{table*}
\begin{table*}[ht]
	\caption{Comparisons with state-of-the-art 4-shot AD methods on industrial domain.}
	\centering
	\label{tab:4s-industial}
    \vspace{-8pt}
	\renewcommand{\arraystretch}{1.0}
	\resizebox{1.0\textwidth}{!}
	{
		\begin{tabular}
        {
        >{\centering\arraybackslash}m{1.0cm}
        >{\centering\arraybackslash}m{2.3cm} 
        *{4}{>{\centering\arraybackslash}p{2.3cm}}
        >{\columncolor{lightgreen}\centering\arraybackslash}m{2.3cm}
        >{\columncolor{lightgreen}\centering\arraybackslash}m{2.3cm}
        >{\columncolor{lightgreen}\centering\arraybackslash}m{2.3cm}
        }
		\toprule
	\multirow{2}{*}{Metric}   &\multirow{2}{*}{Dataset}    &WinCLIP+\scriptsize{\cite{cvpr2023winclip}}    &InCtrl\scriptsize{\cite{cvpr2024inctrl}} &AnomalyCLIP+\scriptsize{\cite{iclr2024anomalyclip}}           &AdaptCLIP\scriptsize{\cite{adaptclip}}                & \textbf{\method}$^\dagger$      & \textbf{\method}$^\ddagger$ & \textbf{\method}$^\star$           \\ 
    &    &\green{\xmark}  &\green{\xmark} &\green{\xmark} &\green{\xmark}  &\red{\cmark} &\red{\cmark} &\red{\cmark} \\
    \midrule
\multirow{7}{*}{\rotatebox[origin=c]{90}{%
    \makecell[c]{Image-Level \\ (AUROC, AUPR)}%
}}

&MVTec	&\pmerror{95.3}{0.1}	, \pmerror{97.7}{0.0}	&\pmerror{93.1}{0.7}	, \pmerror{96.3}{0.5}	&\pmerror{96.1}{0.1}	, \pmerror{97.8}{0.0}	&\pmerror{96.6}{0.3}	, \pmerror{98.4}{0.2}	&\pmerror{97.7}{0.3}	, \pmerror{98.6}{0.2}	&\pmerror{98.7}{0.1}	, \pmerror{99.2}{0.1}	&\pmerror{98.2}{0.1}	, \pmerror{98.8}{0.1}\\
&VisA	&\pmerror{84.3}{0.6}	, \pmerror{85.5}{0.9}	&\pmerror{87.8}{0.2}	, \pmerror{88.0}{0.3}	&\pmerror{88.8}{0.5}	, \pmerror{90.1}{0.7}	&\pmerror{93.1}{0.2}	, \pmerror{94.3}{0.2}	&\pmerror{93.3}{0.2}	, \pmerror{93.8}{0.1}	&\pmerror{96.9}{0.3}	, \pmerror{96.8}{0.3}	&\pmerror{97.1}{0.1}	, \pmerror{96.6}{0.1}\\
&BTAD	&\pmerror{87.8}{0.8}	, \pmerror{88.1}{1.4}	&\pmerror{67.5}{2.4}	, \pmerror{80.9}{1.7}	&\pmerror{90.5}{1.2}	, \pmerror{77.5}{3.1}	&\pmerror{93.3}{0.3}	, \pmerror{96.4}{0.1}	&\pmerror{96.7}{0.2}	, \pmerror{97.8}{0.4}	&\pmerror{96.9}{0.3}	, \pmerror{98.4}{0.3}	&\pmerror{97.5}{0.3}	, \pmerror{98.8}{0.2}\\
&DTD	&\pmerror{98.2}{0.0}	, \pmerror{99.2}{0.1}	&\pmerror{97.7}{0.1}	, \pmerror{98.3}{0.4}	&\pmerror{98.4}{0.1}	, \pmerror{99.4}{0.1}	&\pmerror{98.5}{0.1}	, \pmerror{99.3}{0.0}	&\pmerror{98.4}{0.4}	, \pmerror{99.2}{0.2}	&\pmerror{99.7}{0.1}	, \pmerror{99.7}{0.0}	&\pmerror{99.7}{0.1}	, \pmerror{99.7}{0.1}\\
&KSDD	&\pmerror{94.0}{0.2}	, \pmerror{84.9}{0.5}	&\pmerror{91.6}{0.9}	, \pmerror{84.6}{1.7}	&\pmerror{97.8}{0.1}	, \pmerror{95.0}{0.2}	&\pmerror{97.0}{0.2}	, \pmerror{91.7}{0.9}	&\pmerror{97.3}{0.8}	, \pmerror{91.3}{1.8}	&\pmerror{95.3}{0.6}	, \pmerror{71.7}{1.9}	&\pmerror{96.7}{0.2}	, \pmerror{86.5}{1.7}\\
&RealIAD	&\pmerror{77.0}{0.0}	, \pmerror{73.6}{0.1}	&\pmerror{81.8}{0.0}	, \pmerror{75.6}{0.0}	&\pmerror{78.4}{0.0}	, \pmerror{77.1}{0.0}	&\pmerror{83.9}{0.2}	, \pmerror{82.6}{0.0}	&\pmerror{84.3}{0.2}	, \pmerror{82.6}{0.2}	&\pmerror{90.3}{0.2}	, \pmerror{88.4}{0.3}	&\pmerror{88.5}{0.2}	, \pmerror{87.0}{0.2}\\
\cmidrule{2-9}
&\textbf{Mean}	&89.4	, 88.2	&86.6	, 87.3	&91.7	, 89.5	&93.7	, 93.8	&\sbest{94.6}	, \sbest{93.9}	&\best{96.3}	, 92.4	&\best{96.3}	, \best{94.6}\\
\midrule
\multirow{7}{*}{\rotatebox[origin=c]{90}{%
    \makecell[c]{Pixel-Level \\ (AUROC, AUPR)}%
}}

&MVTec	&\pmerror{94.2}{0.2}	, \pmerror{41.2}{0.9}	&\pmerror{95.8}{0.2}	, \pmerror{50.9}{0.3}	&\pmerror{93.2}{0.1}	, \pmerror{42.4}{0.0}	&\pmerror{94.8}{0.1}	, \pmerror{57.2}{0.8}	&\pmerror{96.3}{0.1}	, \pmerror{58.8}{0.5}	&\pmerror{97.7}{0.1}	, \pmerror{65.4}{0.2}	&\pmerror{97.9}{0.0}	, \pmerror{69.0}{0.4}\\
&VisA	&\pmerror{95.1}{0.2}	, \pmerror{18.1}{1.3}	&\pmerror{90.2}{0.2}	, \pmerror{19.2}{0.6}	&\pmerror{96.9}{0.1}	, \pmerror{27.5}{1.1}	&\pmerror{97.3}{0.0}	, \pmerror{41.8}{0.6}	&\pmerror{98.0}{0.1}	, \pmerror{36.7}{1.3}	&\pmerror{98.6}{0.1}	, \pmerror{45.2}{0.6}	&\pmerror{98.4}{0.1}	, \pmerror{45.5}{0.6}\\
&BTAD	&\pmerror{95.9}{0.1}	, \pmerror{44.0}{0.4}	&\pmerror{96.8}{0.0}	, \pmerror{44.0}{0.2}	&\pmerror{95.7}{0.1}	, \pmerror{45.8}{3.0}	&\pmerror{96.8}{0.0}	, \pmerror{62.3}{0.3}	&\pmerror{96.6}{0.1}	, \pmerror{56.5}{1.4}	&\pmerror{98.4}{0.1}	, \pmerror{73.9}{0.5}	&\pmerror{98.4}{0.0}	, \pmerror{72.7}{1.0}\\
&DTD	&\pmerror{96.8}{0.1}	, \pmerror{49.3}{0.1}	&\pmerror{98.7}{0.1}	, \pmerror{64.9}{0.3}	&\pmerror{97.8}{0.0}	, \pmerror{68.5}{0.2}	&\pmerror{97.8}{0.1}	, \pmerror{78.2}{0.2}	&\pmerror{97.7}{0.1}	, \pmerror{75.2}{0.3}	&\pmerror{99.2}{0.1}	, \pmerror{83.8}{0.1}	&\pmerror{99.3}{0.1}	, \pmerror{86.7}{0.1}\\
&KSDD	&\pmerror{97.5}{0.2}	, \pmerror{19.1}{0.7}	&\pmerror{97.5}{0.3}	, \pmerror{26.0}{1.4}	&\pmerror{98.6}{0.1}	, \pmerror{46.4}{0.7}	&\pmerror{98.0}{0.1}	, \pmerror{56.4}{1.4}	&\pmerror{98.1}{0.2}	, \pmerror{42.5}{5.7}	&\pmerror{99.4}{0.2}	, \pmerror{35.2}{5.3}	&\pmerror{99.3}{0.2}	, \pmerror{44.0}{5.7}\\
&RealIAD	&\pmerror{95.5}{0.0}	, \pmerror{15.4}{0.2}	&\pmerror{96.4}{0.0}	, \pmerror{21.0}{0.0}	&\pmerror{96.7}{0.0}	, \pmerror{28.2}{0.0}	&\pmerror{97.4}{0.0}	, \pmerror{39.1}{0.3}	&\pmerror{98.3}{0.1}	, \pmerror{37.2}{0.2}	&\pmerror{99.1}{0.1}	, \pmerror{48.5}{1.2}	&\pmerror{98.9}{0.1}	, \pmerror{49.8}{0.2}\\

\cmidrule{2-9}
&\textbf{Mean}	&95.8	, 31.2	&95.9	, 37.7	&96.5	, 43.1	&97.0	, \sbest{55.8}	&\sbest{97.5}	, 51.2	&\best{98.7}	, \best{58.7}	&\best{98.7}	, \best{61.3}\\
    \bottomrule
	\end{tabular}
	}
\vspace{-5pt}
\end{table*}

\section{Implementation Details}

\noindent\textbf{Diverse Datasets:} To validate the effectiveness of our \textbf{\method}, we conduct comprehensive experiments on 14 public anomaly detection benchmarks covering two domains, industrial and medical. Following previous works, we use the industrial VisA~\cite{eccv2022visa} as an auxiliary training set to train our \textbf{\method}, and then evaluate its zero-shot or few-shot generalization on 5 industrial benchmarks, MVTec~\cite{cvpr2019mvtec}, BTAD~\cite{isie2021btad}, DTD~\cite{wacv2023dtd}, KSDD~\cite{jim2020ksdd}, and Real-IAD~\cite{cvpr2024realiad}, and 8 medical benchmarks, HeadCT~\cite{salehi2021multiresolution}, BrainMRI~\cite{brainmri}, Br35H~\cite{Br35h}, ISIC~\cite{isic}, ColonDB~\cite{colond}, ClinicDB~\cite{clinicdb}, Kvasir~\cite{icmm2020kvasir}, and Endo~\cite{jhe2017endo}. For the evaluation of VisA~\cite{eccv2022visa}, we use the industrial MVTec~\cite{cvpr2019mvtec} as the auxiliary training set. The statistical information of all these datasets is reported in Tab.~\ref{tab:datasets}.

It should be noted that Real-IAD~\cite{cvpr2024realiad} is the largest anomaly detection dataset, consisting of diverse categories (30 objects) and large-scale images (150k) in industrial domain. In medical domain, either image-level annotations or pixel-level annotations are only provided. Therefore, we only report image anomaly classification performance on HeadCT~\cite{salehi2021multiresolution}, BrainMRI~\cite{brainmri} and Br35H~\cite{Br35h}, and provide pixel-level anomaly segmentation performance on ISIC~\cite{isic}, ColonDB~\cite{colond}, ClinicDB~\cite{clinicdb}, Kvasir~\cite{icmm2020kvasir}, and Endo~\cite{jhe2017endo}. 
Furthermore, considering the lack of normal training images in the medical domain and the fact that objects in this field are typically not rigid like those in the industrial domain, making them unsuitable for a few-shot setting, we only report zero-shot results on the medical domain.

\noindent\textbf{Training Details:} We evaluate a variety of foundational models, such as visual-language models, CLIP (ViT-L/14@336)~\cite{clip}, and self-supervised pure vision models, DINOv2R (ViT-L/14)~\cite{dinov2r} and DINOv3 (ViT-L/16)~\cite{dinov3}, due to the generality of our method.
We extract local global image tokens and local patch tokens from multiple blocks $\{12, 15, 18, 21, 24\}$. When learning the decoupling weights, we apply CAA augmentation to training images with a probability of 0.5 and randomly sample grids from the set $\{2\times2, 3\times3\}$ with equal probability.
At training and testing stages, we resize images to 518$\times$518 and 512$\times$512 for ViT-L/14-based and ViT-L/16-based foundational models, respectively. We train the model for 15 epochs with a learning rate of 0.001, and report the experimental results based on the 15th model. All experiments are conducted using PyTorch with a single NVIDIA H20 GPU.

\section{Competing Methods}
We compare state-of-the-art zero-shot AD methods, such as WinCLIP~\cite{cvpr2023winclip}, AdaCLIP~\cite{eccv24adaclip}, AnomalyCLIP~\cite{iclr2024anomalyclip}, Bayes-PFL~\cite{bayes-pfl}, and AdaptCLIP~\cite{adaptclip}, and few-shot AD methods, such as WinCLIP+~\cite{cvpr2023winclip}, InCtrl~\cite{cvpr2024inctrl}, AnomalyCLIP+, AdaptCLIP~\cite{adaptclip}, and UniVAD~\cite{gu2025univad}, with our \textbf{\method}.  
It is worth noting that state-of-the-art UniVAD~\cite{gu2025univad} uses a multiple-model ensemble (\eg, RAM, Grounding DINO, SAM, CLIP, and DINOv2), which results in slow inference speed and high GPU consumption. We only compared it on MVTec, VisA, and Real-IAD datasets. Following AdaptCLIP~\cite{adaptclip}, we also report the few-shot performance of InCtrl~\cite{cvpr2024inctrl} in both image-level and pixel-level, and extend zero-shot AnomalyCLIP~\cite{iclr2024anomalyclip} to few-shot AnomalyCLIP+. For more details, interested readers refer to AdaptCLIP~\cite{adaptclip}.

\section{Compelete Results}

In main paper, we compare state-of-the-art methods with our \textbf{\method} using the averaged results of all categories for each dataset in AUROC and AUPR for image-level anomaly classification and pixel-level anomaly segmentation. Here, we provide more detailed and comprehensive results, including image-level anomaly classification in AUROC, AUPR, and F1$_\text{max}$, and pixel-level anomaly segmentation in AUROC, AUPR, F1$_\text{max}$, and AUPRO for 6 industrial benchmarks, MVTec ( Tabs.~\ref{tab:clip-mvtec},~\ref{tab:dinov2-mvtec},~\ref{tab:dinov3-mvtec}), VisA (Tabs.~\ref{tab:clip-visa},~\ref{tab:dinov2-visa},~\ref{tab:dinov3-visa}), BTAD (Tabs.~\ref{tab:clip-btad},~\ref{tab:dinov2-btad},~\ref{tab:dinov3-btad}), DTD (Tabs.~\ref{tab:clip-dtd},~\ref{tab:dinov2-dtd},~\ref{tab:dinov3-dtd},), KSDD (Tabs.~\ref{tab:clip-ksdd},~\ref{tab:dinov2-ksdd},~\ref{tab:dinov3-ksdd}), and Real-IAD (Tabs.~\ref{tab:clip-real-iad-01},~\ref{tab:clip-real-iad-24},
~\ref{tab:dinov2-real-iad-01},~\ref{tab:dinov2-real-iad-24},
~\ref{tab:dinov3-real-iad-01},~\ref{tab:dinov3-real-iad-24}), and 8 medical benchmarks covering image anomaly classification (HeadCT, BrainMRI, Br35H in Tab.~\ref{tab:medical-cls}) and pixel anomaly segmentation (ISIC, ColonDB, ClinicDB, Endo, and Kvasir in Tab.~\ref{tab:medical-seg}). In addition, due to space limitations, we mainly compare the state-of-the-art 1-shot methods with our \textbf{\method} in the main paper. Here, following previous works, we additionally provide a comparison of 2-shot and 4-shot in Tabs.~\ref{tab:2s-industial} and~\ref{tab:4s-industial}.

\begin{figure*}[htbp]
    \centering
    \begin{subfigure}[b]{0.4\textwidth}
        \centering
        \includegraphics[width=\linewidth, keepaspectratio]{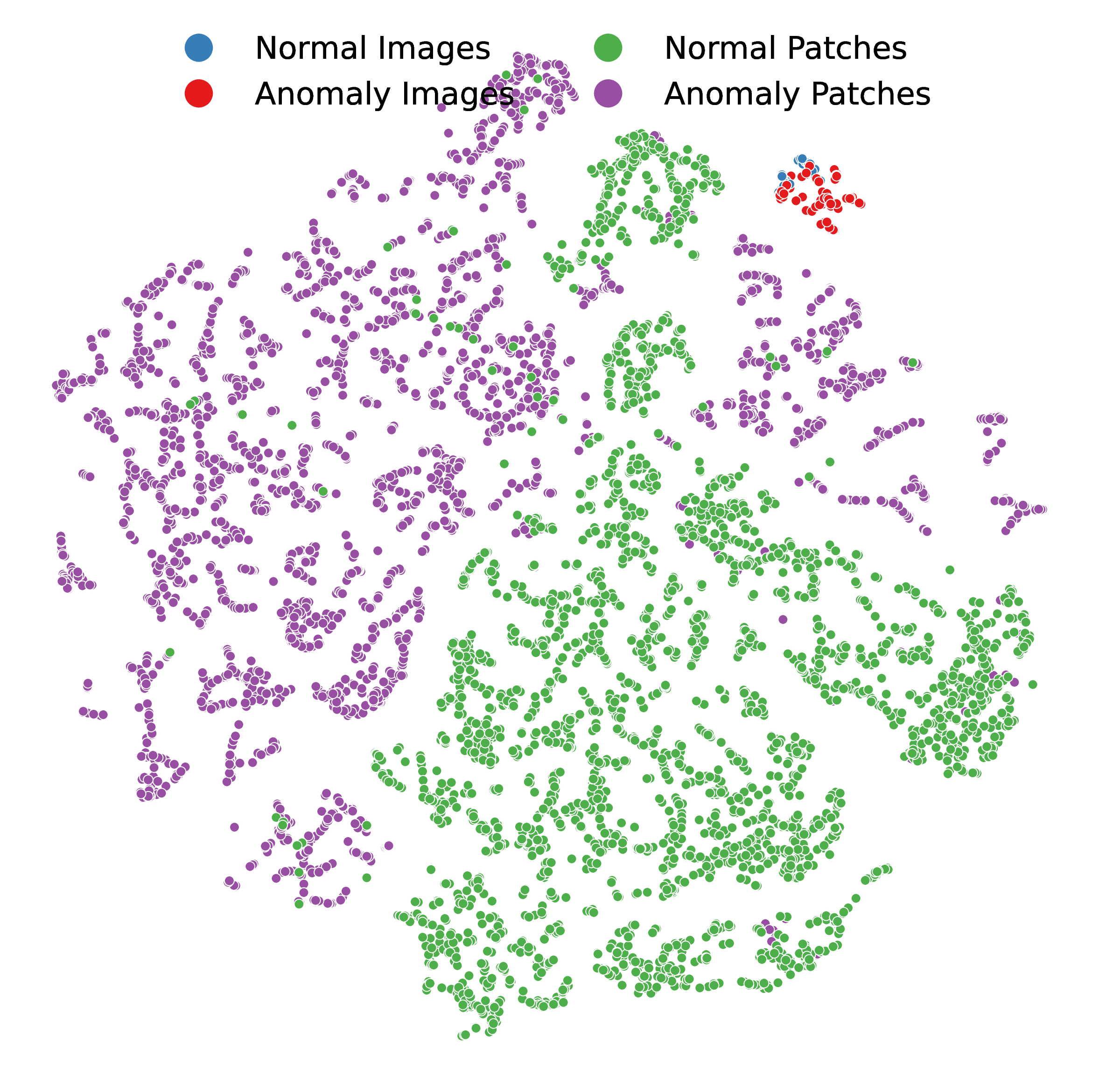}
        \caption{Global vs. Local on DINOv2R ViT-L/14}
    \end{subfigure}
    \hspace{15pt}
    \begin{subfigure}[b]{0.4\textwidth}
        \centering
        \includegraphics[width=\linewidth, keepaspectratio]{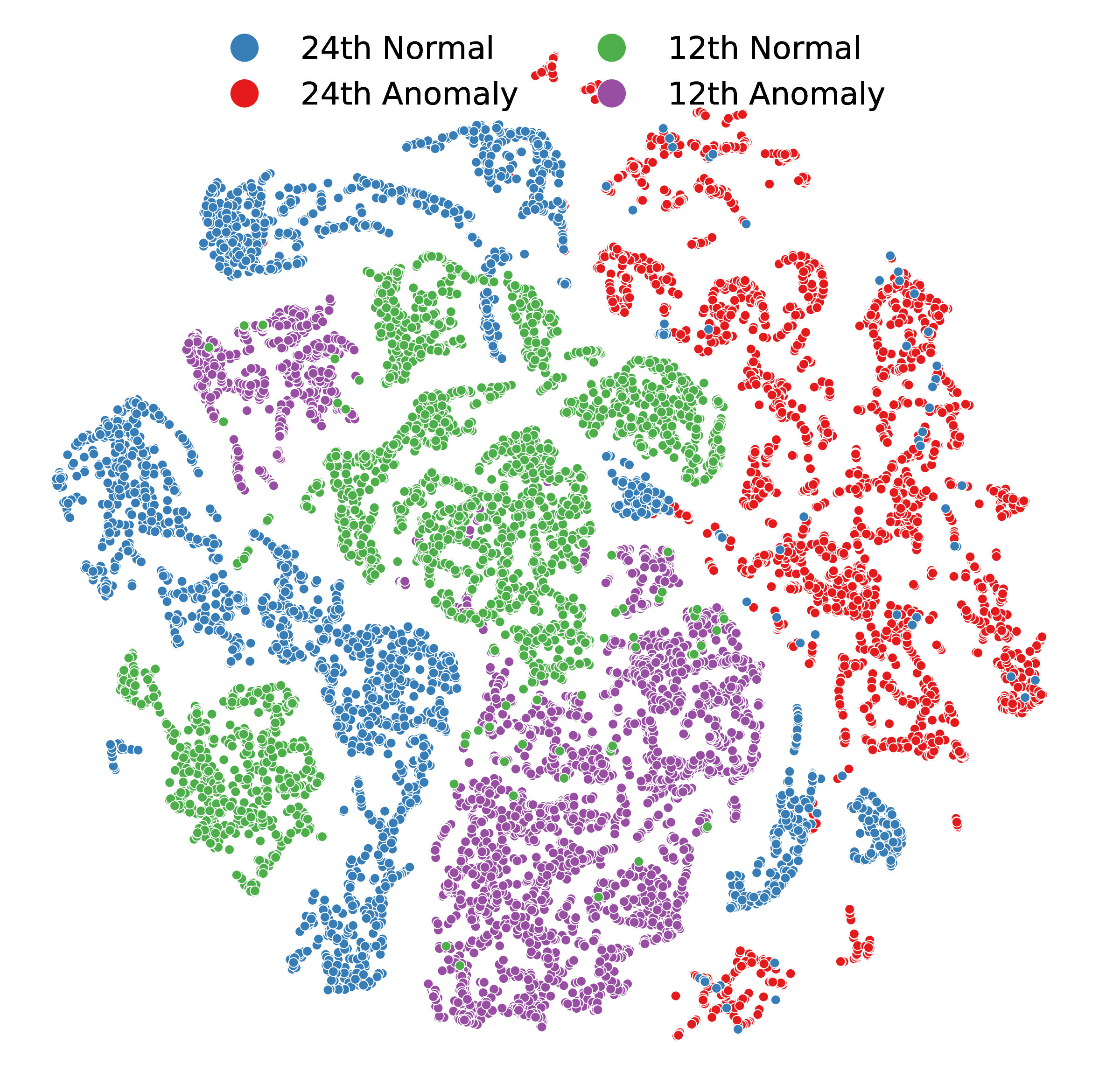}
        \caption{Shallow vs. Deep on DINOv2R ViT-L/14}
    \end{subfigure}
    \begin{subfigure}[b]{0.4\textwidth}
        \centering
        \includegraphics[width=\linewidth, keepaspectratio]{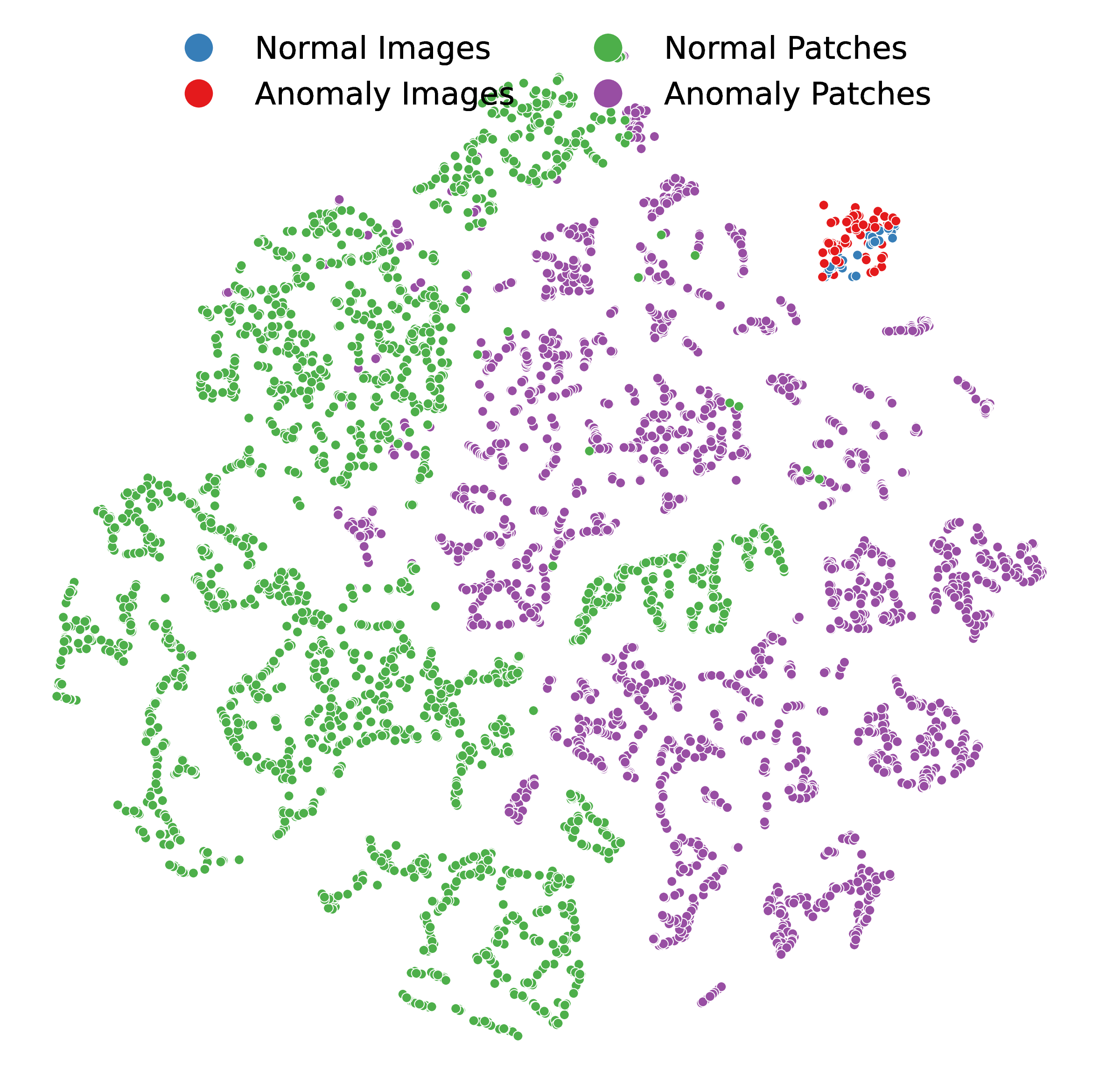}
        \caption{Global Vs. Local on DINOv3 ViT-L/16}
    \end{subfigure}
    \hspace{15pt}
    \begin{subfigure}[b]{0.4\textwidth}
        \centering
        \includegraphics[width=\linewidth, keepaspectratio]{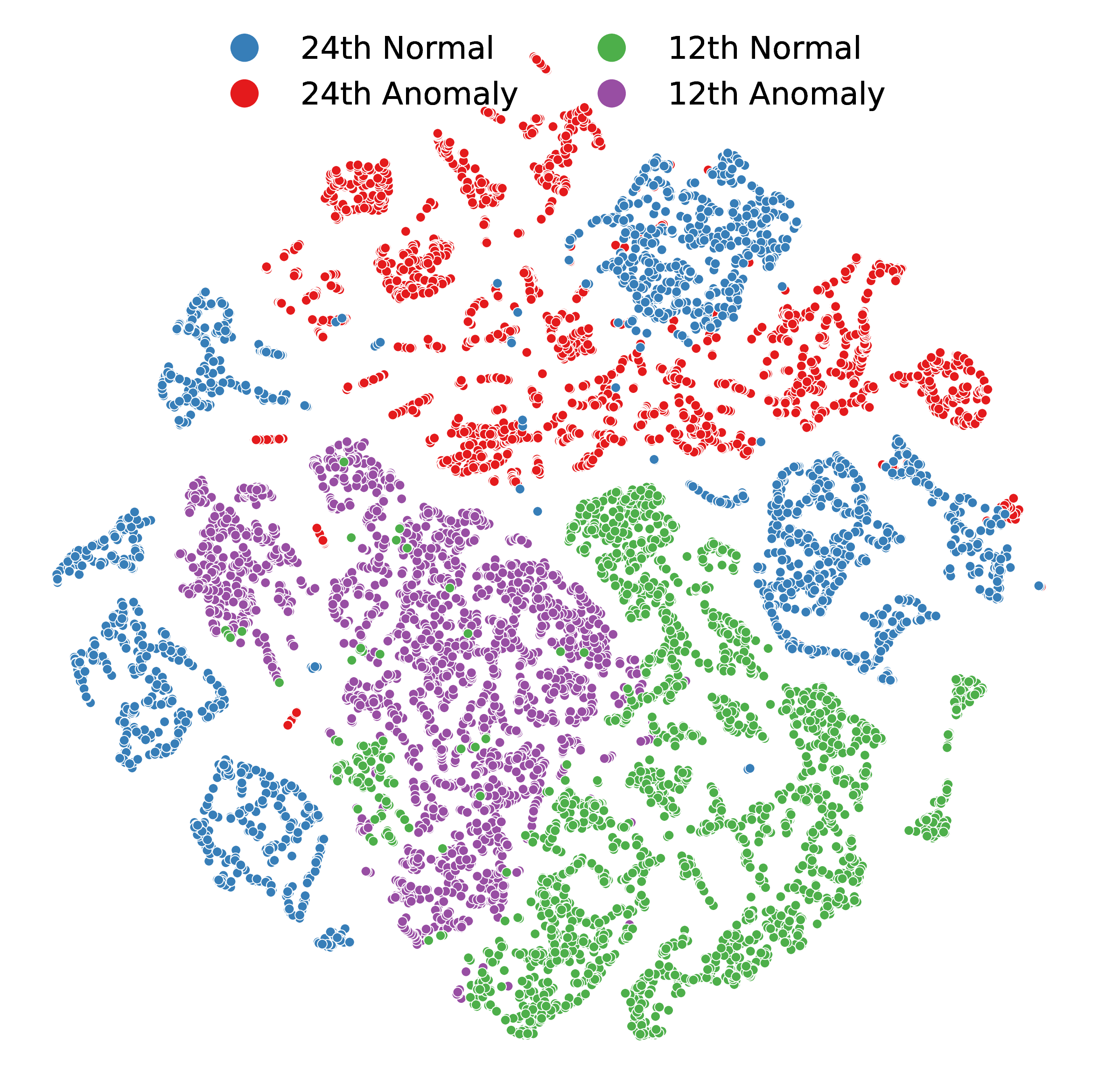}
        \caption{Shallow vs. Deep on DINOv3 ViT-L/16}
    \end{subfigure}
    \vspace{-8pt}
    \caption{\small{t-SNE visualization of features extracted diverse foundation models, \eg, DINOv2R ViT-L/14, and DINOv3 ViT-L/16 on MVTec Test Set (Hazelnuts).}}
    \label{fig:manifold}
\end{figure*}

Furthermore, we only visualized low-dimensional embeddings of CLIP ViT-L/14@336px in our main paper. Here, we additionally add t-SNE visualizations of DINOv2R ViT-L/14 and DINOv3 ViT-L/16, as shown in Fig.~\ref{fig:manifold}. We can see that there are different manifolds between global image tokens and local patch tokens extracted from the same block, as well as between local block tokens extracted from different blocks, regardless of whether the feature extractor is a visual-language CLIP model or self-supervised pure vision models, DINOv2R and DINOv3.

\begin{table*}
  \setlength\tabcolsep{12pt}
  \centering
  \small
  \caption{\small{Image anomaly classification and pixel anomaly segmentation results on \textbf{MVTec} with zero-shot and few-shot~\textbf{\method$^\dagger$}.}}\label{tab:clip-mvtec}
  \vspace{-5pt}
  \resizebox{1.0\textwidth}{!}{
\begin{tabular}
     {
        >{\centering\arraybackslash}p{0.5cm}
        >{\centering\arraybackslash}p{1.5cm} 
        *{3}{>{\centering\arraybackslash}p{1.5cm}}
         >{\centering\arraybackslash}p{0.1cm} 
        *{4}{>{\centering\arraybackslash}p{1.5cm}}
        }
    \toprule
     \multirow{2}{*}{\textbf{Shot}}  &\multirow{2}{*}{\textbf{Catergoies}} &\multicolumn{3}{c}{\textbf{Anomaly Classification}} & &\multicolumn{4}{c}{\textbf{Anomaly Segmentation}} \\
     \cline{3-5} \cline{7-10}\\
       & &\textbf{I-AUROC}	&\textbf{I-AUPR}& \textbf{I-$\text{F1}_{\text{max}}$}	&	&\textbf{P-AUROC}	&\textbf{P-AUPR}	&\textbf{P-$\text{F1}_{\text{max}}$}
       &\textbf{P-AUPRO} \\
    \midrule
    \multirow{16}{*}{\textbf{0-shot}}	&carpet	&99.9	&99.8	&99.4	&&99.2	&67.6	&63.6	&98.3\\
	&bottle	&93.7	&98.2	&92.2	&&91.4	&55.5	&52.7	&84.7\\
	&hazelnut	&95.4	&97.2	&93.8	&&96.7	&52.6	&51.5	&93.0\\
	&leather	&100.0	&99.9	&100.0	&&99.4	&53.5	&51.5	&99.1\\
	&cable	&76.7	&82.6	&79.5	&&80.2	&21.2	&27.7	&69.3\\
\bottomrule
  \end{tabular}
  }
  \vspace{-10pt}
\end{table*}

\begin{table*}
  \setlength\tabcolsep{12pt}
  \centering
  \small
  \caption{\small{Image anomaly classification and pixel anomaly segmentation results on \textbf{MVTec} with zero-shot and few-shot~\textbf{\method$^\ddagger$}.}}\label{tab:dinov2-mvtec}
  \vspace{-5pt}
  \resizebox{1.0\textwidth}{!}{
    \begin{tabular}
     {
        >{\centering\arraybackslash}p{0.5cm}
        >{\centering\arraybackslash}p{1.5cm} 
        *{3}{>{\centering\arraybackslash}p{1.5cm}}
         >{\centering\arraybackslash}p{0.001cm} 
        *{4}{>{\centering\arraybackslash}p{1.5cm}}
        }
    \toprule
     \multirow{2}{*}{\textbf{Shot}}  &\multirow{2}{*}{\textbf{Catergoies}} &\multicolumn{3}{c}{\textbf{Anomaly Classification}} & &\multicolumn{4}{c}{\textbf{Anomaly Segmentation}} \\
     \cline{3-5} \cline{7-10}\\
       & &\textbf{I-AUROC}	&\textbf{I-AUPR}& \textbf{I-$\text{F1}_{\text{max}}$}	&	&\textbf{P-AUROC}	&\textbf{P-AUPR}	&\textbf{P-$\text{F1}_{\text{max}}$}
       &\textbf{P-AUPRO} \\
    \midrule
    \multirow{16}{*}{\textbf{0-shot}}	&carpet	&98.7	&99.5	&97.3	&&99.5	&75.9	&70.1	&98.9\\
	&bottle	&99.7	&99.7	&98.4	&&93.9	&68.5	&62.9	&90.0\\
	&hazelnut	&96.5	&98.0	&92.6	&&97.0	&58.1	&55.0	&91.4\\
	&leather	&99.9	&99.8	&98.9	&&99.4	&50.9	&49.4	&99.2\\
	&cable	&81.6	&88.9	&80.4	&&81.5	&23.1	&28.3	&73.7\\
\bottomrule
  \end{tabular}
  }
  \vspace{-10pt}
\end{table*}

\begin{table*}
  \setlength\tabcolsep{12pt}
  \centering
  \small
  \caption{\small{Image anomaly classification and pixel anomaly segmentation results on \textbf{MVTec} with zero-shot and few-shot~\textbf{\method$^\star$}.}}\label{tab:dinov3-mvtec}
  \vspace{-5pt}
  \resizebox{1.0\textwidth}{!}{
    \begin{tabular}
     {
        >{\centering\arraybackslash}p{0.5cm}
        >{\centering\arraybackslash}p{1.5cm} 
        *{3}{>{\centering\arraybackslash}p{1.5cm}}
         >{\centering\arraybackslash}p{0.1cm} 
        *{4}{>{\centering\arraybackslash}p{1.5cm}}
        }
    \toprule
     \multirow{2}{*}{\textbf{Shot}}  &\multirow{2}{*}{\textbf{Catergoies}} &\multicolumn{3}{c}{\textbf{Anomaly Classification}} & &\multicolumn{4}{c}{\textbf{Anomaly Segmentation}} \\
     \cline{3-5} \cline{7-10}\\
       & &\textbf{I-AUROC}	&\textbf{I-AUPR}& \textbf{I-$\text{F1}_{\text{max}}$}	&	&\textbf{P-AUROC}	&\textbf{P-AUPR}	&\textbf{P-$\text{F1}_{\text{max}}$}
       &\textbf{P-AUPRO} \\
    \midrule
    \multirow{16}{*}{\textbf{0-shot}}	&carpet	&99.8	&99.8	&98.9	&&99.4	&76.1	&70.4	&98.9\\	
	&bottle	&98.3	&99.5	&97.6	&&95.8	&73.2	&66.0	&92.7\\	
	&hazelnut	&97.7	&98.6	&95.6	&&97.4	&62.0	&58.5	&91.7\\	
	&leather	&100.0	&99.9	&100.0	&&99.5	&58.0	&52.9	&99.4\\	
	&cable	&81.4	&88.0	&82.8	&&81.0	&17.8	&24.7	&72.7\\
\bottomrule
  \end{tabular}
  }
  \vspace{-10pt}
\end{table*}

\begin{table*}
  \setlength\tabcolsep{8pt}
  \centering
  \small
  \caption{\small{Image anomaly classification and pixel anomaly segmentation results on \textbf{VisA} with zero-shot and few-shot~\textbf{\method$^\dagger$}.}}\label{tab:clip-visa}
  \vspace{-5pt}
  \resizebox{1.0\textwidth}{!}{
    \begin{tabular}
    {
        >{\centering\arraybackslash}p{0.5cm}
        >{\centering\arraybackslash}p{1.5cm} 
        *{3}{>{\centering\arraybackslash}p{1.5cm}}
         >{\centering\arraybackslash}p{0.1cm} 
        *{4}{>{\centering\arraybackslash}p{1.5cm}}
        }
    \toprule
    \multirow{2}{*}{\textbf{Shot}}  &\multirow{2}{*}{\textbf{Catergoies}} &\multicolumn{3}{c}{\textbf{Anomaly Classification}} & &\multicolumn{4}{c}{\textbf{Anomaly Segmentation}} \\
     \cline{3-5} \cline{7-10}\\
       & &\textbf{I-AUROC}	&\textbf{I-AUPR}& \textbf{I-$\text{F1}_{\text{max}}$}	&	&\textbf{P-AUROC}	&\textbf{P-AUPR}	&\textbf{P-$\text{F1}_{\text{max}}$} &\textbf{P-AUPRO}	\\
    \midrule
    \multirow{13}{*}{\textbf{0-shot}}	&candle	&94.5	&95.0	&86.9	&&99.3	&39.5	&46.9	&95.6\\
	&capsules	&94.3	&96.6	&91.5	&&99.0	&51.4	&53.1	&94.8\\
	&cashew	&89.4	&95.4	&86.0	&&90.3	&16.3	&21.8	&92.8\\
	&chewinggum	&97.3	&98.9	&96.4	&&99.6	&83.2	&76.6	&95.3\\
	&fryum	&97.0	&98.5	&93.1	&&94.7	&24.2	&31.1	&93.0\\
\bottomrule
  \end{tabular}
  }
  \vspace{-10pt}
\end{table*}

\begin{table*}
  \setlength\tabcolsep{8pt}
  \centering
  \small
  \caption{\small{Image anomaly classification and pixel anomaly segmentation results on \textbf{VisA} with zero-shot and few-shot~\textbf{\method$^\ddagger$}.}}\label{tab:dinov2-visa}
  \vspace{-5pt}
  \resizebox{1.0\textwidth}{!}{
    \begin{tabular}
     {
        >{\centering\arraybackslash}p{0.5cm}
        >{\centering\arraybackslash}p{1.5cm} 
        *{3}{>{\centering\arraybackslash}p{1.5cm}}
         >{\centering\arraybackslash}p{0.1cm} 
        *{4}{>{\centering\arraybackslash}p{1.5cm}}
        }
    \toprule
     \multirow{2}{*}{\textbf{Shot}}  &\multirow{2}{*}{\textbf{Catergoies}} &\multicolumn{3}{c}{\textbf{Anomaly Classification}} & &\multicolumn{4}{c}{\textbf{Anomaly Segmentation}} \\
     \cline{3-5} \cline{7-10}\\
       & &\textbf{I-AUROC}	&\textbf{I-AUPR}& \textbf{I-$\text{F1}_{\text{max}}$}	&	&\textbf{P-AUROC}	&\textbf{P-AUPR}	&\textbf{P-$\text{F1}_{\text{max}}$}
       &\textbf{P-AUPRO} \\
    \midrule
    \multirow{13}{*}{\textbf{0-shot}}	&candle	&94.7	&95.7	&88.9	&&99.5	&51.8	&54.8	&93.0\\	
	&capsules	&98.7	&99.2	&96.4	&&99.6	&73.3	&68.9	&98.8\\	
	&cashew	&91.3	&95.5	&89.8	&&90.8	&16.9	&19.6	&92.9\\	
	&chewinggum	&99.4	&99.6	&97.5	&&99.7	&87.3	&79.3	&95.3\\	
	&fryum	&97.2	&98.6	&94.7	&&94.2	&26.1	&30.7	&93.8\\
\bottomrule
  \end{tabular}
  }
  \vspace{-10pt}
\end{table*}

\begin{table*}
  \setlength\tabcolsep{8pt}
  \centering
  \small
  \caption{\small{Image anomaly classification and pixel anomaly segmentation results on \textbf{VisA} with zero-shot and few-shot~\textbf{\method$^\star$}.}}\label{tab:dinov3-visa}
  \vspace{-5pt}
  \resizebox{1.0\textwidth}{!}{
    \begin{tabular}
     {
        >{\centering\arraybackslash}p{0.5cm}
        >{\centering\arraybackslash}p{1.5cm} 
        *{3}{>{\centering\arraybackslash}p{1.5cm}}
         >{\centering\arraybackslash}p{0.1cm} 
        *{4}{>{\centering\arraybackslash}p{1.5cm}}
        }
    \toprule
     \multirow{2}{*}{\textbf{Shot}}  &\multirow{2}{*}{\textbf{Catergoies}} &\multicolumn{3}{c}{\textbf{Anomaly Classification}} & &\multicolumn{4}{c}{\textbf{Anomaly Segmentation}} \\
     \cline{3-5} \cline{7-10}\\
       & &\textbf{I-AUROC}	&\textbf{I-AUPR}& \textbf{I-$\text{F1}_{\text{max}}$}	&	&\textbf{P-AUROC}	&\textbf{P-AUPR}	&\textbf{P-$\text{F1}_{\text{max}}$}
       &\textbf{P-AUPRO} \\
    \midrule

    \multirow{13}{*}{\textbf{0-shot}}	&candle	&94.9	&95.8	&89.8	&&99.4	&52.6	&54.0	&92.9\\
	&capsules	&98.7	&99.4	&98.0	&&99.6	&76.8	&72.0	&99.3\\
	&cashew	&95.1	&97.6	&93.1	&&93.4	&20.1	&22.9	&95.4\\
	&chewinggum	&98.9	&99.4	&96.9	&&99.6	&88.2	&81.3	&94.5\\
	&fryum	&95.6	&97.9	&94.0	&&95.1	&30.1	&35.5	&92.9\\
\bottomrule
  \end{tabular}
  }
  \vspace{-10pt}
\end{table*}

\begin{table*}
  \setlength\tabcolsep{12pt}
  \centering
  \small
  \caption{\small{Image anomaly classification and pixel anomaly segmentation results on \textbf{BTAD} with zero-shot and few-shot~\textbf{\method$^\dagger$}.}}\label{tab:clip-btad}
  \vspace{-5pt}
  \resizebox{1.0\textwidth}{!}{
    \begin{tabular}{
        >{\centering\arraybackslash}p{0.5cm}
        >{\centering\arraybackslash}p{1.5cm} 
        *{3}{>{\centering\arraybackslash}p{1.5cm}}
         >{\centering\arraybackslash}p{0.1cm} 
        *{4}{>{\centering\arraybackslash}p{1.5cm}}
        }
    \toprule
    \multirow{2}{*}{\textbf{Shot}}  &\multirow{2}{*}{\textbf{Catergoies}} &\multicolumn{3}{c}{\textbf{Anomaly Classification}} & &\multicolumn{4}{c}{\textbf{Anomaly Segmentation}} \\
     \cline{3-5} \cline{7-10}\\
       & &\textbf{I-AUROC}	&\textbf{I-AUPR}& \textbf{I-$\text{F1}_{\text{max}}$}	&	&\textbf{P-AUROC}	&\textbf{P-AUPR}	&\textbf{P-$\text{F1}_{\text{max}}$} &\textbf{P-AUPRO}	\\
 \midrule
 \multirow{4}{*}{\textbf{0-shot}}	&1	&98.1	&99.2	&97.0	&&95.2	&52.8	&57.4	&82.6\\	
	&2	&91.6	&98.7	&95.0	&&94.9	&61.3	&61.4	&69.2\\	
	&3	&99.6	&95.6	&89.4	&&97.1	&14.4	&25.7	&91.9\\	
\rowcolor{lightgreen}	&\textbf{Mean}	&96.4	&97.8	&93.8	&&95.7	&42.8	&48.2	&81.3\\	
\midrule
\multirow{4}{*}{\textbf{1-shot}}	&1	&\pmerror{98.4}{0.2}	&\pmerror{99.4}{0.0}	&\pmerror{98.0}{0.0}	&&\pmerror{97.3}{0.1}	&\pmerror{63.7}{0.2}	&\pmerror{64.4}{0.1}	&\pmerror{85.6}{0.6}\\
\bottomrule
  \end{tabular}
  }
  \vspace{-5pt}
\end{table*}

\begin{table*}
  \setlength\tabcolsep{12pt}
  \centering
  \small
  \caption{\small{Image anomaly classification and pixel anomaly segmentation results on \textbf{BTAD} with zero-shot and few-shot~\textbf{\method$^\ddagger$}.}}\label{tab:dinov2-btad}
  \vspace{-5pt}
  \resizebox{1.0\textwidth}{!}{
    \begin{tabular}
     {
        >{\centering\arraybackslash}p{0.5cm}
        >{\centering\arraybackslash}p{1.5cm} 
        *{3}{>{\centering\arraybackslash}p{1.5cm}}
         >{\centering\arraybackslash}p{0.1cm} 
        *{4}{>{\centering\arraybackslash}p{1.5cm}}
        }
    \toprule
     \multirow{2}{*}{\textbf{Shot}}  &\multirow{2}{*}{\textbf{Catergoies}} &\multicolumn{3}{c}{\textbf{Anomaly Classification}} & &\multicolumn{4}{c}{\textbf{Anomaly Segmentation}} \\
     \cline{3-5} \cline{7-10}\\
       & &\textbf{I-AUROC}	&\textbf{I-AUPR}& \textbf{I-$\text{F1}_{\text{max}}$}	&	&\textbf{P-AUROC}	&\textbf{P-AUPR}	&\textbf{P-$\text{F1}_{\text{max}}$}
       &\textbf{P-AUPRO} \\
    \midrule
\multirow{4}{*}{\textbf{0-shot}}	&1	&94.9	&98.3	&94.6	&&97.4	&65.2	&63.7	&85.1\\
	&2	&90.1	&98.4	&95.2	&&96.9	&74.6	&68.9	&75.7\\
	&3	&98.1	&92.4	&89.2	&&99.0	&42.0	&45.0	&97.4\\
\rowcolor{lightgreen}	&\textbf{Mean}	&94.4	&96.3	&93.0	&&97.7	&60.6	&59.2	&86.1\\
\midrule    \multirow{4}{*}{\textbf{1-shot}}	&1	&\pmerror{98.7}{0.3}	&\pmerror{99.3}{0.1}	&\pmerror{97.2}{1.2}	&&\pmerror{98.2}{0.1}	&\pmerror{68.7}{0.9}	&\pmerror{66.2}{0.7}	&\pmerror{86.0}{0.5}\\
\bottomrule
  \end{tabular}
  }
  \vspace{-5pt}
\end{table*}

\begin{table*}
  \setlength\tabcolsep{12pt}
  \centering
  \small
  \caption{\small{Image anomaly classification and pixel anomaly segmentation results on \textbf{BTAD} with zero-shot and few-shot~\textbf{\method$^\star$}.}}\label{tab:dinov3-btad}
  \vspace{-5pt}
  \resizebox{1.0\textwidth}{!}{
    \begin{tabular}
     {
        >{\centering\arraybackslash}p{0.5cm}
        >{\centering\arraybackslash}p{1.5cm} 
        *{3}{>{\centering\arraybackslash}p{1.5cm}}
         >{\centering\arraybackslash}p{0.1cm} 
        *{4}{>{\centering\arraybackslash}p{1.5cm}}
        }
    \toprule
     \multirow{2}{*}{\textbf{Shot}}  &\multirow{2}{*}{\textbf{Catergoies}} &\multicolumn{3}{c}{\textbf{Anomaly Classification}} & &\multicolumn{4}{c}{\textbf{Anomaly Segmentation}} \\
     \cline{3-5} \cline{7-10}\\
       & &\textbf{I-AUROC}	&\textbf{I-AUPR}& \textbf{I-$\text{F1}_{\text{max}}$}	&	&\textbf{P-AUROC}	&\textbf{P-AUPR}	&\textbf{P-$\text{F1}_{\text{max}}$}
       &\textbf{P-AUPRO} \\
    \midrule
\multirow{4}{*}{\textbf{0-shot}}	&1	&99.9	&99.7	&99.0	&&97.9	&68.6	&64.8	&90.0\\
	&2	&85.0	&97.5	&93.4	&&96.8	&75.4	&69.6	&77.4\\
	&3	&99.2	&93.8	&86.8	&&98.7	&32.5	&39.1	&96.0\\
\rowcolor{lightgreen}	&\textbf{Mean}	&94.7	&97.0	&93.1	&&97.8	&58.9	&57.8	&87.8\\
\midrule   \multirow{4}{*}{\textbf{1-shot}}	&1	&\pmerror{100.0}{0.1}	&\pmerror{99.7}{0.0}	&\pmerror{99.7}{0.6}	&&\pmerror{98.0}{0.1}	&\pmerror{68.5}{0.6}	&\pmerror{66.1}{0.7}	&\pmerror{88.8}{0.5}\\
\bottomrule
  \end{tabular}
  }
  \vspace{-10pt}
\end{table*}

\begin{table*}
  \setlength\tabcolsep{8pt}
  \centering
  \small
  \caption{\small{Image anomaly classification and pixel anomaly segmentation results on \textbf{DTD} with zero-shot and few-shot~\textbf{\method$^\dagger$}.}}\label{tab:clip-dtd}
  \vspace{-5pt}
  \resizebox{1.0\textwidth}{!}{
    \begin{tabular}
    {
        >{\centering\arraybackslash}p{0.5cm}
        >{\centering\arraybackslash}p{1.5cm} 
        *{3}{>{\centering\arraybackslash}p{1.5cm}}
         >{\centering\arraybackslash}p{0.1cm} 
        *{4}{>{\centering\arraybackslash}p{1.5cm}}
        }
    \toprule
    \multirow{2}{*}{\textbf{Shot}}  &\multirow{2}{*}{\textbf{Catergoies}} &\multicolumn{3}{c}{\textbf{Anomaly Classification}} & &\multicolumn{4}{c}{\textbf{Anomaly Segmentation}} \\
     \cline{3-5} \cline{7-10}\\
       & &\textbf{I-AUROC}	&\textbf{I-AUPR}& \textbf{I-$\text{F1}_{\text{max}}$}	&	&\textbf{P-AUROC}	&\textbf{P-AUPR}	&\textbf{P-$\text{F1}_{\text{max}}$} &\textbf{P-AUPRO}	\\
    \midrule\multirow{13}{*}{\textbf{0-shot}}	&Woven\_001	&100.0	&99.8	&100.0	&&99.8	&75.8	&69.6	&99.5\\
	&Woven\_127	&95.9	&97.1	&92.7	&&95.1	&54.1	&55.3	&92.4\\
	&Woven\_104	&99.1	&99.7	&98.8	&&97.6	&69.3	&64.4	&95.8\\
	&Stratified\_154	&99.1	&99.8	&99.4	&&99.2	&68.4	&65.7	&99.2\\
	&Blotchy\_099	&100.0	&99.9	&100.0	&&98.8	&59.3	&56.5	&95.0\\
\bottomrule
  \end{tabular}
  }
  \vspace{-10pt}
\end{table*}

\begin{table*}
  \setlength\tabcolsep{8pt}
  \centering
  \small
  \caption{\small{Image anomaly classification and pixel anomaly segmentation results on \textbf{DTD} with zero-shot and few-shot~\textbf{\method$^\ddagger$}.}}\label{tab:dinov2-dtd}
  \vspace{-5pt}
  \resizebox{1.0\textwidth}{!}{
    \begin{tabular}
     {
        >{\centering\arraybackslash}p{0.5cm}
        >{\centering\arraybackslash}p{1.5cm} 
        *{3}{>{\centering\arraybackslash}p{1.5cm}}
         >{\centering\arraybackslash}p{0.1cm} 
        *{4}{>{\centering\arraybackslash}p{1.5cm}}
        }
    \toprule
     \multirow{2}{*}{\textbf{Shot}}  &\multirow{2}{*}{\textbf{Catergoies}} &\multicolumn{3}{c}{\textbf{Anomaly Classification}} & &\multicolumn{4}{c}{\textbf{Anomaly Segmentation}} \\
     \cline{3-5} \cline{7-10}\\
       & &\textbf{I-AUROC}	&\textbf{I-AUPR}& \textbf{I-$\text{F1}_{\text{max}}$}	&	&\textbf{P-AUROC}	&\textbf{P-AUPR}	&\textbf{P-$\text{F1}_{\text{max}}$}
       &\textbf{P-AUPRO} \\
\midrule    \multirow{13}{*}{\textbf{0-shot}}	&Woven\_001	&100.0	&99.8	&100.0	&&99.9	&84.5	&75.4	&99.6\\
	&Woven\_127	&95.7	&96.8	&91.9	&&95.8	&69.4	&66.8	&95.7\\
	&Woven\_104	&99.8	&99.8	&99.4	&&99.4	&78.7	&71.6	&98.7\\
	&Stratified\_154	&97.4	&99.2	&96.4	&&99.6	&79.6	&74.8	&99.3\\
	&Blotchy\_099	&76.9	&94.0	&88.9	&&95.5	&54.1	&53.4	&76.3\\
\bottomrule
  \end{tabular}
  }
  \vspace{-10pt}
\end{table*}

\begin{table*}
  \setlength\tabcolsep{8pt}
  \centering
  \small
  \caption{\small{Image anomaly classification and pixel anomaly segmentation results on \textbf{DTD} with zero-shot and few-shot~\textbf{\method$^\star$}.}}\label{tab:dinov3-dtd}
  \vspace{-5pt}
  \resizebox{1.0\textwidth}{!}{
    \begin{tabular}
     {
        >{\centering\arraybackslash}p{0.5cm}
        >{\centering\arraybackslash}p{1.5cm} 
        *{3}{>{\centering\arraybackslash}p{1.5cm}}
         >{\centering\arraybackslash}p{0.1cm} 
        *{4}{>{\centering\arraybackslash}p{1.5cm}}
        }
    \toprule
     \multirow{2}{*}{\textbf{Shot}}  &\multirow{2}{*}{\textbf{Catergoies}} &\multicolumn{3}{c}{\textbf{Anomaly Classification}} & &\multicolumn{4}{c}{\textbf{Anomaly Segmentation}} \\
     \cline{3-5} \cline{7-10}\\
       & &\textbf{I-AUROC}	&\textbf{I-AUPR}& \textbf{I-$\text{F1}_{\text{max}}$}	&	&\textbf{P-AUROC}	&\textbf{P-AUPR}	&\textbf{P-$\text{F1}_{\text{max}}$}
       &\textbf{P-AUPRO} \\
\midrule   \multirow{13}{*}{\textbf{0-shot}}	&Woven\_001	&100.0	&99.8	&100.0	&&99.9	&85.2	&75.6	&99.6\\
	&Woven\_127	&96.5	&97.5	&92.6	&&97.8	&73.1	&68.3	&96.6\\
	&Woven\_104	&99.7	&99.8	&98.7	&&99.7	&84.7	&76.8	&99.2\\
	&Stratified\_154	&97.8	&99.5	&97.5	&&99.4	&80.1	&73.0	&99.4\\
	&Blotchy\_099	&83.1	&95.6	&88.9	&&98.3	&61.5	&57.4	&90.5\\
\bottomrule
  \end{tabular}
  }
  \vspace{-10pt}
\end{table*}

\begin{table*}
  \setlength\tabcolsep{8pt}
  \centering
  \small
  \caption{\small{Image anomaly classification and pixel anomaly segmentation results on \textbf{KSDD} with zero-shot and few-shot~\textbf{\method$^\dagger$}.}}\label{tab:clip-ksdd}
  \vspace{-5pt}
  \resizebox{1.0\textwidth}{!}{
    \begin{tabular}
    {
        >{\centering\arraybackslash}p{1cm}
        >{\centering\arraybackslash}p{2.5cm} 
        *{3}{>{\centering\arraybackslash}p{1.5cm}}
         >{\centering\arraybackslash}p{0.1cm} 
        *{4}{>{\centering\arraybackslash}p{1.5cm}}
        }
    \toprule
    \multirow{2}{*}{\textbf{Shot}}  &\multirow{2}{*}{\textbf{Catergoies}} &\multicolumn{3}{c}{\textbf{Anomaly Classification}} & &\multicolumn{4}{c}{\textbf{Anomaly Segmentation}} \\
     \cline{3-5} \cline{7-10}\\
       & &\textbf{I-AUROC}	&\textbf{I-AUPR}& \textbf{I-$\text{F1}_{\text{max}}$}	&	&\textbf{P-AUROC}	&\textbf{P-AUPR}	&\textbf{P-$\text{F1}_{\text{max}}$} &\textbf{P-AUPRO}	\\
    \midrule

{\textbf{0-shot}}	&electrical\_commutators	&96.3	&83.4	&80.0	&&98.1	&31.8	&43.6	&86.3\\
{\textbf{1-shot}}	&electrical\_commutators	&\pmerror{97.7}{0.7}	&\pmerror{92.4}{3.2}	&\pmerror{89.7}{2.7}	&&\pmerror{98.0}{0.0}	&\pmerror{45.0}{5.8}	&\pmerror{50.9}{3.9}	&\pmerror{87.0}{0.1}\\
{\textbf{2-shot}}	&electrical\_commutators	&\pmerror{97.7}{0.6}	&\pmerror{93.3}{1.3}	&\pmerror{89.4}{3.6}	&&\pmerror{98.1}{0.2}	&\pmerror{46.1}{3.9}	&\pmerror{50.6}{3.6}	&\pmerror{87.4}{0.6}\\
{\textbf{4-shot}}	&electrical\_commutators	&\pmerror{97.3}{0.8}	&\pmerror{91.3}{1.8}	&\pmerror{87.1}{3.6}	&&\pmerror{98.1}{0.2}	&\pmerror{42.5}{5.7}	&\pmerror{47.9}{5.3}	&\pmerror{87.4}{0.7}\\
    
    \bottomrule
  \end{tabular}
  }
  \vspace{-10pt}
\end{table*}

\begin{table*}
  \setlength\tabcolsep{8pt}
  \centering
  \small
  \caption{\small{Image anomaly classification and pixel anomaly segmentation results on \textbf{KSDD} with zero-shot and few-shot~\textbf{\method$^\ddagger$}.}}\label{tab:dinov2-ksdd}
  \vspace{-5pt}
  \resizebox{1.0\textwidth}{!}{
    \begin{tabular}
     {
        >{\centering\arraybackslash}p{1cm}
        >{\centering\arraybackslash}p{2.5cm} 
        *{3}{>{\centering\arraybackslash}p{1.5cm}}
         >{\centering\arraybackslash}p{0.1cm} 
        *{4}{>{\centering\arraybackslash}p{1.5cm}}
        }
    \toprule
     \multirow{2}{*}{\textbf{Shot}}  &\multirow{2}{*}{\textbf{Catergoies}} &\multicolumn{3}{c}{\textbf{Anomaly Classification}} & &\multicolumn{4}{c}{\textbf{Anomaly Segmentation}} \\
     \cline{3-5} \cline{7-10}\\
       & &\textbf{I-AUROC}	&\textbf{I-AUPR}& \textbf{I-$\text{F1}_{\text{max}}$}	&	&\textbf{P-AUROC}	&\textbf{P-AUPR}	&\textbf{P-$\text{F1}_{\text{max}}$}
       &\textbf{P-AUPRO} \\
    \midrule
{\textbf{0-shot}}	&electrical\_commutators	&91.6	&59.5	&73.0	&&98.4	&15.8	&24.5	&90.2\\
{\textbf{1-shot}}	&electrical\_commutators	&\pmerror{94.7}{0.4}	&\pmerror{68.7}{2.0}	&\pmerror{78.7}{1.5}	&&\pmerror{99.4}{0.3}	&\pmerror{34.9}{2.7}	&\pmerror{46.1}{0.9}	&\pmerror{97.7}{0.6}\\
{\textbf{2-shot}}	&electrical\_commutators	&\pmerror{95.3}{0.2}	&\pmerror{71.2}{1.2}	&\pmerror{78.8}{1.4}	&&\pmerror{99.5}{0.1}	&\pmerror{36.9}{2.7}	&\pmerror{47.7}{1.6}	&\pmerror{97.6}{0.3}\\
{\textbf{4-shot}}	&electrical\_commutators	&\pmerror{95.3}{0.6}	&\pmerror{71.7}{1.9}	&\pmerror{79.2}{1.1}	&&\pmerror{99.4}{0.2}	&\pmerror{35.2}{5.3}	&\pmerror{45.6}{5.1}	&\pmerror{96.8}{1.5}\\   

    \bottomrule
  \end{tabular}
  }
  \vspace{-10pt}
\end{table*}

\begin{table*}
  \setlength\tabcolsep{8pt}
  \centering
  \small
  \caption{\small{Image anomaly classification and pixel anomaly segmentation results on \textbf{KSDD} with zero-shot and few-shot~\textbf{\method$^\star$}.}}\label{tab:dinov3-ksdd}
  \vspace{-5pt}
  \resizebox{1.0\textwidth}{!}{
    \begin{tabular}
     {
        >{\centering\arraybackslash}p{1cm}
        >{\centering\arraybackslash}p{2.5cm} 
        *{3}{>{\centering\arraybackslash}p{1.5cm}}
         >{\centering\arraybackslash}p{0.1cm} 
        *{4}{>{\centering\arraybackslash}p{1.5cm}}
        }
    \toprule
     \multirow{2}{*}{\textbf{Shot}}  &\multirow{2}{*}{\textbf{Catergoies}} &\multicolumn{3}{c}{\textbf{Anomaly Classification}} & &\multicolumn{4}{c}{\textbf{Anomaly Segmentation}} \\
     \cline{3-5} \cline{7-10}\\
       & &\textbf{I-AUROC}	&\textbf{I-AUPR}& \textbf{I-$\text{F1}_{\text{max}}$}	&	&\textbf{P-AUROC}	&\textbf{P-AUPR}	&\textbf{P-$\text{F1}_{\text{max}}$}
       &\textbf{P-AUPRO} \\
    \midrule
{\textbf{0-shot}}	&electrical\_commutators	&91.0	&64.5	&67.8	&&99.0	&22.6	&32.6	&92.0\\	
{\textbf{1-shot}}	&electrical\_commutators	&\pmerror{96.9}{0.3}	&\pmerror{84.7}{2.0}	&\pmerror{83.4}{0.9}	&&\pmerror{99.3}{0.1}	&\pmerror{45.2}{3.7}	&\pmerror{50.4}{2.4}	&\pmerror{94.7}{0.3}\\	
{\textbf{2-shot}}	&electrical\_commutators	&\pmerror{97.2}{0.4}	&\pmerror{87.5}{0.9}	&\pmerror{84.0}{1.3}	&&\pmerror{99.3}{0.2}	&\pmerror{45.0}{2.2}	&\pmerror{49.9}{1.2}	&\pmerror{94.6}{0.4}\\	
{\textbf{4-shot}}	&electrical\_commutators	&\pmerror{96.7}{0.2}	&\pmerror{86.5}{1.7}	&\pmerror{82.4}{1.4}	&&\pmerror{99.3}{0.2}	&\pmerror{44.0}{5.7}	&\pmerror{49.3}{4.1}	&\pmerror{93.9}{2.3}\\	   

    \bottomrule
  \end{tabular}
  }
  \vspace{-10pt}
\end{table*}

\begin{table*}
  \setlength\tabcolsep{15pt}
  \centering
  \small
  \caption{\small{Image anomaly classification results (I-AUROC, I-AUPR, I-$\text{F1}_{\text{max}}$) on \textbf{Medical} with zero-shot~\textbf{\method}.}}\label{tab:medical-cls}
  \vspace{-5pt}
  \resizebox{1.0\textwidth}{!}{
    \begin{tabular}
     {
        >{\centering\arraybackslash}p{1.5cm} 
        *{3}{>{\centering\arraybackslash}p{4.5cm}}
        }
    \toprule
    \textbf{Catergoies} &\textbf{\method$^\dagger$}   &\textbf{\method$^\dagger$}    &\textbf{\method$^\star$} \\
\midrule
    HeadCT	&97.0\quad ,  \quad  96.7\quad ,  \quad  91.8	&95.2\quad ,  \quad  94.9\quad ,  \quad  89.2	&95.7\quad ,  \quad  94.8\quad ,  \quad  90.1 \\
BrainMRI	&95.7\quad ,  \quad  96.4\quad ,  \quad  95.1	&91.7\quad ,  \quad  94.8\quad ,  \quad  88.0	&91.9\quad ,  \quad  95.0\quad ,  \quad  88.3 \\
Br35H	&97.4\quad ,  \quad  97.1\quad ,  \quad  93.0	&92.5\quad ,  \quad  93.9\quad ,  \quad  85.7	&92.3\quad ,  \quad  93.8\quad ,  \quad  85.3 \\
\rowcolor{lightgreen} \textbf{Mean}	&96.7\quad ,  \quad  96.7\quad ,  \quad  93.3	&93.1\quad ,  \quad  94.5\quad ,  \quad  87.7	&93.3\quad ,  \quad  94.5\quad ,  \quad  87.9 \\    
\bottomrule
  \end{tabular}
  }
  \vspace{-10pt}
\end{table*}

\begin{table*}
  \setlength\tabcolsep{8pt}
  \centering
  \small
  \caption{\small{Pixel anomaly segmentation results (P-AUROC, P-AUPR, P-$\text{F1}_{\text{max}}$, P-AUPRO) on \textbf{Medical} with zero-shot~\textbf{\method$^\star$}.}}\label{tab:medical-seg}
  \vspace{-5pt}
  \resizebox{1.0\textwidth}{!}{
    \begin{tabular}
     {
        >{\centering\arraybackslash}p{1.5cm} 
        *{3}{>{\centering\arraybackslash}p{5.5cm}}
        }
    \toprule
    \textbf{Catergoies} &\textbf{\method$^\dagger$}   &\textbf{\method$^\dagger$}    &\textbf{\method$^\star$} \\
\midrule
ISIC	&92.6\quad , \quad 84.2\quad , \quad 76.8\quad , \quad 87.8	&95.0\quad , \quad 88.5\quad , \quad 81.4\quad ,  \quad  90.4	&95.9\quad ,  \quad  90.6\quad ,  \quad  83.6\quad ,  \quad  92.6 \\
ColonDB	&84.0\quad ,  \quad  36.2\quad ,  \quad  40.8\quad ,  \quad  79.2	&86.0\quad ,  \quad  35.5\quad ,  \quad  43.1\quad ,  \quad  79.8	&86.7\quad ,  \quad  33.3\quad ,  \quad  43.2\quad ,  \quad  80.6 \\
ClinicDB	&90.1\quad ,  \quad  55.8\quad ,  \quad  55.6\quad ,  \quad  81.1	&91.3\quad ,  \quad  59.3\quad ,  \quad  60.6\quad ,  \quad  82.2	&91.5\quad ,  \quad  57.5\quad ,  \quad  59.5\quad ,  \quad  83.4 \\
Endo	&90.7\quad ,  \quad  62.5\quad ,  \quad  61.5\quad ,  \quad  79.3	&92.6\quad ,  \quad  67.3\quad ,  \quad  66.4\quad ,  \quad  82.7	&92.8\quad ,  \quad  68.5\quad ,  \quad  66.1\quad ,  \quad  83.3 \\
Kvasir	&87.5\quad ,  \quad  57.5\quad ,  \quad  58.6\quad ,  \quad  59.7	&91.1\quad ,  \quad  65.2\quad ,  \quad  65.3\quad ,  \quad  53.1	&91.7\quad ,  \quad  65.9\quad ,  \quad  65.8\quad ,  \quad  54.9 \\
\rowcolor{lightgreen} \textbf{Mean}	&89.0\quad ,  \quad  59.2\quad ,  \quad  58.7\quad ,  \quad  77.4	&91.2\quad ,  \quad  63.2\quad ,  \quad  63.4\quad ,  \quad  77.6	&91.7\quad ,  \quad  63.2\quad ,  \quad  63.6\quad ,  \quad  79.0 \\
\bottomrule
  \end{tabular}
  }
  \vspace{-10pt}
\end{table*}

\begin{table*}
  \setlength\tabcolsep{12pt}
  \centering
  \small
  \caption{\small{Image anomaly classification and pixel anomaly segmentation results on \textbf{Real-IAD} with zero-shot and 1-shot~\textbf{\method$^\dagger$}.}}\label{tab:clip-real-iad-01}
  \vspace{-5pt}
  \resizebox{1.0\textwidth}{!}{
    \begin{tabular}
    {
        >{\centering\arraybackslash}p{0.5cm}
        >{\centering\arraybackslash}p{1.5cm} 
        *{3}{>{\centering\arraybackslash}p{1.5cm}}
         >{\centering\arraybackslash}p{0.1cm} 
        *{4}{>{\centering\arraybackslash}p{1.5cm}}
        }
    \toprule
    \multirow{2}{*}{\textbf{Shot}}  &\multirow{2}{*}{\textbf{Catergoies}} &\multicolumn{3}{c}{\textbf{Anomaly Classification}} & &\multicolumn{4}{c}{\textbf{Anomaly Segmentation}} \\
     \cline{3-5} \cline{7-10}\\
       & &\textbf{I-AUROC}	&\textbf{I-AUPR}& \textbf{I-$\text{F1}_{\text{max}}$}	&	&\textbf{P-AUROC}	&\textbf{P-AUPR}	&\textbf{P-$\text{F1}_{\text{max}}$} &\textbf{P-AUPRO}	\\
    \midrule
\multirow{31}{*}{\textbf{0-shot}}	&audiojack	&67.2	&56.7	&53.5	&&97.6	&27.7	&37.6	&84.3\\
	&bottle\_cap	&64.7	&55.0	&62.8	&&99.1	&15.7	&21.5	&95.4\\
	&button\_battery	&81.5	&86.3	&78.1	&&98.0	&45.0	&46.5	&89.4\\
	&end\_cap	&71.6	&78.7	&74.5	&&96.2	&22.3	&30.3	&88.4\\
	&eraser	&89.9	&89.3	&79.5	&&99.8	&44.4	&48.7	&98.2\\
\bottomrule
  \end{tabular}
  }
  \vspace{-10pt}
\end{table*}

\begin{table*}
  \setlength\tabcolsep{12pt}
  \centering
  \small
  \caption{\small{Image anomaly classification and pixel anomaly segmentation results on \textbf{Real-IAD} with 2-shot and 4-shot~\textbf{\method$^\dagger$}.}}\label{tab:clip-real-iad-24}
  \vspace{-5pt}
  \resizebox{1.0\textwidth}{!}{
    \begin{tabular}
    {
        >{\centering\arraybackslash}p{0.5cm}
        >{\centering\arraybackslash}p{1.5cm} 
        *{3}{>{\centering\arraybackslash}p{1.5cm}}
         >{\centering\arraybackslash}p{0.1cm} 
        *{4}{>{\centering\arraybackslash}p{1.5cm}}
        }
    \toprule
    \multirow{2}{*}{\textbf{Shot}}  &\multirow{2}{*}{\textbf{Catergoies}} &\multicolumn{3}{c}{\textbf{Anomaly Classification}} & &\multicolumn{4}{c}{\textbf{Anomaly Segmentation}} \\
     \cline{3-5} \cline{7-10}\\
       & &\textbf{I-AUROC}	&\textbf{I-AUPR}& \textbf{I-$\text{F1}_{\text{max}}$}	&	&\textbf{P-AUROC}	&\textbf{P-AUPR}	&\textbf{P-$\text{F1}_{\text{max}}$} &\textbf{P-AUPRO}	\\
    \midrule
\multirow{31}{*}{\textbf{2-shot}}	&audiojack	&\pmerror{77.1}{0.6}	&\pmerror{71.6}{0.5}	&\pmerror{61.0}{0.7}	&&\pmerror{98.5}{0.1}	&\pmerror{30.5}{1.2}	&\pmerror{42.2}{0.8}	&\pmerror{90.8}{0.4}\\
	&bottle\_cap	&\pmerror{82.3}{0.1}	&\pmerror{77.6}{0.2}	&\pmerror{77.1}{0.3}	&&\pmerror{99.7}{0.1}	&\pmerror{35.7}{0.4}	&\pmerror{39.9}{0.2}	&\pmerror{97.4}{0.1}\\
	&button\_battery	&\pmerror{79.0}{1.0}	&\pmerror{85.5}{0.8}	&\pmerror{75.7}{0.9}	&&\pmerror{97.4}{0.2}	&\pmerror{48.0}{3.4}	&\pmerror{48.9}{3.3}	&\pmerror{86.3}{1.4}\\
	&end\_cap	&\pmerror{76.0}{1.4}	&\pmerror{81.6}{1.3}	&\pmerror{76.1}{0.2}	&&\pmerror{98.7}{0.0}	&\pmerror{25.1}{0.8}	&\pmerror{31.9}{1.1}	&\pmerror{95.1}{0.2}\\
	&eraser	&\pmerror{93.5}{0.5}	&\pmerror{92.4}{0.3}	&\pmerror{83.8}{0.3}	&&\pmerror{99.8}{0.0}	&\pmerror{40.8}{1.5}	&\pmerror{44.3}{0.6}	&\pmerror{98.7}{0.0}\\
\bottomrule
  \end{tabular}
  }
  \vspace{-10pt}
\end{table*}

\begin{table*}
  \setlength\tabcolsep{12pt}
  \centering
  \small
  \caption{\small{Image anomaly classification and pixel anomaly segmentation results on \textbf{Real-IAD} with zero-shot and 1-shot~\textbf{\method$^\ddagger$}.}}\label{tab:dinov2-real-iad-01}
  \vspace{-5pt}
  \resizebox{1.0\textwidth}{!}{
    \begin{tabular}
     {
        >{\centering\arraybackslash}p{0.5cm}
        >{\centering\arraybackslash}p{1.5cm} 
        *{3}{>{\centering\arraybackslash}p{1.5cm}}
         >{\centering\arraybackslash}p{0.1cm} 
        *{4}{>{\centering\arraybackslash}p{1.5cm}}
        }
    \toprule
     \multirow{2}{*}{\textbf{Shot}}  &\multirow{2}{*}{\textbf{Catergoies}} &\multicolumn{3}{c}{\textbf{Anomaly Classification}} & &\multicolumn{4}{c}{\textbf{Anomaly Segmentation}} \\
     \cline{3-5} \cline{7-10}\\
       & &\textbf{I-AUROC}	&\textbf{I-AUPR}& \textbf{I-$\text{F1}_{\text{max}}$}	&	&\textbf{P-AUROC}	&\textbf{P-AUPR}	&\textbf{P-$\text{F1}_{\text{max}}$}
       &\textbf{P-AUPRO} \\
    \midrule
 \multirow{31}{*}{\textbf{0-shot}}	&audiojack	&70.4	&62.6	&55.6	&&97.8	&34.5	&39.9	&86.0\\	
	&bottle\_cap	&82.5	&79.6	&72.5	&&99.7	&30.7	&35.2	&98.0\\	
	&button\_battery	&76.9	&84.3	&74.0	&&98.2	&43.5	&43.3	&92.0\\	
	&end\_cap	&71.9	&78.6	&74.9	&&98.0	&25.2	&31.7	&93.6\\	
	&eraser	&92.2	&90.8	&81.4	&&99.9	&56.2	&55.1	&99.1\\
\bottomrule
  \end{tabular}
  }
  \vspace{-10pt}
\end{table*}

\begin{table*}
  \setlength\tabcolsep{12pt}
  \centering
  \small
  \caption{\small{Image anomaly classification and pixel anomaly segmentation results on \textbf{Real-IAD} with 2-shot and 4-shot~\textbf{\method$^\ddagger$}.}}\label{tab:dinov2-real-iad-24}
  \vspace{-5pt}
  \resizebox{1.0\textwidth}{!}{
    \begin{tabular}
     {
        >{\centering\arraybackslash}p{0.5cm}
        >{\centering\arraybackslash}p{1.5cm} 
        *{3}{>{\centering\arraybackslash}p{1.5cm}}
         >{\centering\arraybackslash}p{0.1cm} 
        *{4}{>{\centering\arraybackslash}p{1.5cm}}
        }
    \toprule
     \multirow{2}{*}{\textbf{Shot}}  &\multirow{2}{*}{\textbf{Catergoies}} &\multicolumn{3}{c}{\textbf{Anomaly Classification}} & &\multicolumn{4}{c}{\textbf{Anomaly Segmentation}} \\
     \cline{3-5} \cline{7-10}\\
       & &\textbf{I-AUROC}	&\textbf{I-AUPR}& \textbf{I-$\text{F1}_{\text{max}}$}	&	&\textbf{P-AUROC}	&\textbf{P-AUPR}	&\textbf{P-$\text{F1}_{\text{max}}$}
       &\textbf{P-AUPRO} \\
    \midrule
  \multirow{31}{*}{\textbf{2-shot}}	&audiojack	&\pmerror{86.4}{1.3}	&\pmerror{81.9}{1.9}	&\pmerror{71.1}{1.8}	&&\pmerror{99.4}{0.0}	&\pmerror{53.0}{0.9}	&\pmerror{55.0}{0.1}	&\pmerror{95.6}{0.2}\\	
	&bottle\_cap	&\pmerror{96.4}{0.5}	&\pmerror{95.8}{0.6}	&\pmerror{89.6}{1.1}	&&\pmerror{99.9}{0.0}	&\pmerror{43.1}{0.1}	&\pmerror{46.7}{0.0}	&\pmerror{98.8}{0.0}\\	
	&button\_battery	&\pmerror{87.2}{0.9}	&\pmerror{90.5}{0.6}	&\pmerror{82.2}{1.1}	&&\pmerror{98.5}{0.1}	&\pmerror{54.1}{1.4}	&\pmerror{51.4}{0.8}	&\pmerror{92.1}{1.1}\\	
	&end\_cap	&\pmerror{87.8}{1.3}	&\pmerror{89.7}{0.7}	&\pmerror{83.4}{1.5}	&&\pmerror{99.4}{0.0}	&\pmerror{27.0}{0.1}	&\pmerror{34.4}{0.0}	&\pmerror{97.6}{0.2}\\	
	&eraser	&\pmerror{95.2}{0.4}	&\pmerror{94.1}{0.7}	&\pmerror{86.2}{1.1}	&&\pmerror{99.8}{0.1}	&\pmerror{52.2}{0.6}	&\pmerror{51.7}{0.0}	&\pmerror{99.0}{0.3}\\
\bottomrule
  \end{tabular}
  }
  \vspace{-10pt}
\end{table*}

\begin{table*}
  \setlength\tabcolsep{12pt}
  \centering
  \small
  \caption{\small{Image anomaly classification and pixel anomaly segmentation results on \textbf{Real-IAD} with zero-shot and 1-shot~\textbf{\method$^\star$}.}}\label{tab:dinov3-real-iad-01}
  \vspace{-5pt}
  \resizebox{1.0\textwidth}{!}{
    \begin{tabular}
     {
        >{\centering\arraybackslash}p{0.5cm}
        >{\centering\arraybackslash}p{1.5cm} 
        *{3}{>{\centering\arraybackslash}p{1.5cm}}
         >{\centering\arraybackslash}p{0.1cm} 
        *{4}{>{\centering\arraybackslash}p{1.5cm}}
        }
    \toprule
     \multirow{2}{*}{\textbf{Shot}}  &\multirow{2}{*}{\textbf{Catergoies}} &\multicolumn{3}{c}{\textbf{Anomaly Classification}} & &\multicolumn{4}{c}{\textbf{Anomaly Segmentation}} \\
     \cline{3-5} \cline{7-10}\\
       & &\textbf{I-AUROC}	&\textbf{I-AUPR}& \textbf{I-$\text{F1}_{\text{max}}$}	&	&\textbf{P-AUROC}	&\textbf{P-AUPR}	&\textbf{P-$\text{F1}_{\text{max}}$}
       &\textbf{P-AUPRO} \\
    \midrule
  \multirow{31}{*}{\textbf{0-shot}}	&audiojack	&68.1	&57.4	&54.4	&&97.7	&19.5	&24.8	&83.8\\	
	&bottle\_cap	&74.9	&73.3	&65.9	&&99.4	&26.2	&30.7	&96.7\\	
	&button\_battery	&74.0	&81.9	&73.7	&&97.9	&42.8	&43.7	&88.1\\	
	&end\_cap	&71.7	&78.7	&74.3	&&96.7	&22.9	&31.7	&88.7\\	
	&eraser	&89.1	&87.8	&77.4	&&99.8	&55.5	&54.5	&98.1\\
\bottomrule
  \end{tabular}
  }
  \vspace{-10pt}
\end{table*}

\begin{table*}
  \setlength\tabcolsep{12pt}
  \centering
  \small
  \caption{\small{Image anomaly classification and pixel anomaly segmentation results on \textbf{Real-IAD} with 2-shot and 4-shot~\textbf{\method$^\star$}.}}\label{tab:dinov3-real-iad-24}
  \vspace{-5pt}
  \resizebox{1.0\textwidth}{!}{
    \begin{tabular}
     {
        >{\centering\arraybackslash}p{0.5cm}
        >{\centering\arraybackslash}p{1.5cm} 
        *{3}{>{\centering\arraybackslash}p{1.5cm}}
         >{\centering\arraybackslash}p{0.1cm} 
        *{4}{>{\centering\arraybackslash}p{1.5cm}}
        }
    \toprule
     \multirow{2}{*}{\textbf{Shot}}  &\multirow{2}{*}{\textbf{Catergoies}} &\multicolumn{3}{c}{\textbf{Anomaly Classification}} & &\multicolumn{4}{c}{\textbf{Anomaly Segmentation}} \\
     \cline{3-5} \cline{7-10}\\
       & &\textbf{I-AUROC}	&\textbf{I-AUPR}& \textbf{I-$\text{F1}_{\text{max}}$}	&	&\textbf{P-AUROC}	&\textbf{P-AUPR}	&\textbf{P-$\text{F1}_{\text{max}}$}
       &\textbf{P-AUPRO} \\
    \midrule
   \multirow{31}{*}{\textbf{2-shot}}	&audiojack	&\pmerror{83.9}{0.7}	&\pmerror{80.4}{0.7}	&\pmerror{70.2}{0.8}	&&\pmerror{99.0}{0.1}	&\pmerror{51.6}{0.3}	&\pmerror{56.8}{0.2}	&\pmerror{91.9}{0.5}\\	
	&bottle\_cap	&\pmerror{88.4}{0.4}	&\pmerror{87.2}{0.5}	&\pmerror{76.6}{0.6}	&&\pmerror{99.9}{0.0}	&\pmerror{41.6}{0.3}	&\pmerror{43.5}{0.5}	&\pmerror{98.8}{0.1}\\	
	&button\_battery	&\pmerror{83.6}{0.6}	&\pmerror{88.4}{0.4}	&\pmerror{78.6}{0.3}	&&\pmerror{98.0}{0.1}	&\pmerror{53.4}{1.1}	&\pmerror{52.6}{0.5}	&\pmerror{90.0}{0.7}\\	
	&end\_cap	&\pmerror{82.4}{1.2}	&\pmerror{84.6}{0.8}	&\pmerror{80.2}{0.8}	&&\pmerror{99.1}{0.1}	&\pmerror{22.2}{1.0}	&\pmerror{30.6}{0.8}	&\pmerror{96.4}{0.3}\\	
	&eraser	&\pmerror{93.4}{0.3}	&\pmerror{91.9}{0.7}	&\pmerror{83.8}{0.8}	&&\pmerror{99.8}{0.0}	&\pmerror{52.2}{1.2}	&\pmerror{51.8}{0.6}	&\pmerror{98.9}{0.0}\\
\bottomrule
  \end{tabular}
  }
  \vspace{-10pt}
\end{table*}